\documentclass[runningheads]{llncs}
\usepackage[T1]{fontenc}
\usepackage{graphicx}
\usepackage{setspace}
\usepackage{booktabs}
\usepackage[misc]{ifsym}
\usepackage{amssymb}

\usepackage{xurl}
\usepackage{mwe}

\usepackage{multirow}
\usepackage{mathtools}
\usepackage{algorithm}
\usepackage{algpseudocode}
\usepackage{bbm}

\begin{document}

\title{Unsupervised Continual Clustering via Forward-Backward Knowledge Distillation}
\author{
Mohammadreza Sadeghi \and
Sareh Soleimani \and
Zihan Wang \and
Narges Armanfard
}

\institute{
Department of Electrical and Computer Engineering, McGill University,
Montreal, QC H3A 0E9, Canada, and
Mila -- Quebec AI Institute,
Montreal, QC H2S 3H1, Canada
}

\titlerunning{Unsupervised Continual Clustering (FBCC)}

\maketitle              

\begin{abstract}
Unsupervised Continual Learning (UCL) aims to enable neural networks to learn sequential tasks without labels or access to past data. A major challenge in this setting is Catastrophic Forgetting, where models forget previously learned tasks upon learning new ones. This challenge is amplified in UCL due to the absence of labels to guide learning and memory retention. Existing mitigation strategies, such as knowledge distillation and replay buffers, often raise memory and privacy concerns. Moreover, current UCL methods largely overlook clustering-specific objectives. To fill this gap, we introduce Unsupervised Continual Clustering (UCC) and propose Forward-Backward Knowledge Distillation for Continual Clustering (FBCC). 
FBCC employs a continual teacher network with a clustering projector and lightweight task-specific students. Through a dual-phase forward–backward distillation process, the teacher learns new clusters while preserving previously discovered cluster structure without storing past data. FBCC represents a pioneering approach to UCC, demonstrating improved clustering performance across sequential tasks. 
Experiments on four benchmark datasets demonstrate that FBCC consistently outperforms existing continual learning baselines in clustering accuracy while significantly reducing catastrophic forgetting.

\keywords{Unsupervised continual learning \and Continual clustering \and Lifelong learning \and knowledge distillation \and Incremental learning.}
\end{abstract}
\section{Introduction}
\label{sec:intro}
Continual Learning (CL) \cite{wang2023comprehensive} aims to enable neural networks to learn a sequence of \textit{tasks} without revisiting past data. Each task presents a portion of a dataset, inaccessible in its entirety. 
Existing CL paradigms include supervised continual learning (SCL), semi-supervised continual learning (SeCL), and unsupervised continual learning (UCL). In contrast to SCL and SeCL, UCL operates without labels and focuses on learning stable representations from evolving data streams. Unlike conventional offline learning, where all data are available simultaneously, continual learning must adapt to sequentially arriving tasks while mitigating catastrophic forgetting. For example, in offline unsupervised contrastive learning frameworks, such as \cite{beclr2024}, an encoder is trained with access to all unlabeled data simultaneously to learn general-purpose feature representations. These representations are then leveraged to improve performance on downstream few-shot tasks, where a limited number of labeled samples are available at evaluation time. In contrast, continual learning frameworks such as the one proposed in this paper, do not have access to all past data. Instead, they must adapt to new tasks sequentially, without revisiting old samples, while preserving previously learned representations.

Existing UCL methods primarily focus on learning stable representations over time \cite{lump,cassle,SCALE} and clustering is treated as a downstream procedure applied after representation learning. However, many real-world applications require clustering itself to be the primary objective, where cluster assignments must be discovered and preserved sequentially without access to previous data. 

To address this gap, we introduce Unsupervised Continual Clustering (UCC), a setting in which cluster structure must be jointly learned and optimized alongside representation learning, and maintained across sequential tasks without labels or replay. In UCC, forgetting is reflected not only in representation drift but in degradation of previously discovered cluster assignments. 

The main challenge of all of the CL approaches is ``Catastrophic Forgetting'' (CF) \cite{wang2023comprehensive}. In SCL, SeCL, and UCL, several strategies attempt to mitigate CF. Some use generative replay with discriminator regularization \cite{wang2021ordisco}, but training generative models is resource-intensive. Others store past samples in replay buffers \cite{lump,li2022learning,mai2021supervised}, though privacy concerns often prohibit this. Knowledge distillation (KD) methods \cite{cassle,li2022learning} transfer \textit{insights} from previous tasks by storing models from prior tasks in memory, but this is memory-inefficient, particularly for large networks. Moreover, methods like \cite{cassle} retain only a single past model, which leads to forgetting when faced with many tasks.

To overcome these limitations, we propose an innovative solution termed \textbf{F}orward-\textbf{B}ackward Knowledge Distillation for mitigating CF in the domain of unsupervised \textbf{C}ontinual \textbf{C}lustering (\textbf{FBCC}). In FBCC, we introduce a single continual learner, also referred to as the ``teacher'', which comprises a deep neural network with a high number of parameters, together with a cluster projector that maps outputs into a clustering space. For each task, we train a lightweight student model with significantly fewer parameters to specialize in reproducing the representations learned by the teacher for that task. During subsequent tasks, the teacher is trained on new data while being explicitly regularized by previously learned student models through forward knowledge distillation, which constrains the teacher’s representations to remain close to those of the frozen students on past tasks. In this way, knowledge from earlier tasks is progressively consolidated into the teacher without storing past data. 

 To sum up, the main novelties of this paper are as follows:
\begin{itemize}
    \item  FBCC stands out as the first unsupervised continual clustering framework that jointly integrates representation learning and clustering, in a sequential, no-replay setting.
    \item FBCC mitigates catastrophic forgetting through a dual-phase forward-backward knowledge distillation strategy, where a compact, task-specific student is distilled after each task to guide the teacher in preserving past knowledge while adapting to new tasks. 
    \item FBCC offers a memory-efficient approach to continual clustering, where task-specific knowledge is stored in specialized light-weight student models rather than storing samples or large-scale models from past tasks.
    \item Extensive experiments on four benchmark datasets demonstrate that FBCC consistently outperforms both state-of-the-art unsupervised and supervised continual learning methods in clustering accuracy and forgetting.
\end{itemize}

\section{Related Work}
\label{sec:relatedwork}
\textbf{Unsupervised Continual learning.} The primary challenge in CL lies in combating CF, a phenomenon involving the loss of performance on previously learned tasks when learning new tasks. Traditional CL approaches mitigate CF using regularization-based methods \cite{Jung2020nips,Paik2019OvercomingCF,Aljundi_2018_ECCV}, architectural expansion or parameter isolation strategies \cite{rusu2022progressive,Rypes2024iclr,Wang_2023}, and replay-based mechanisms \cite{arani2022learning,yoon2022online,ashfahani2022}. However, these approaches assume the availability of task- and class-specific labels, which are incompatible with our label-free, no-replay setting. Conversely, unsupervised continual learning, without task and class labels, presents a significantly more complex challenge. Several works employ generative replay or memory-intensive architectures to preserve past representations \cite{RAMAPURAM2020381,Rao2019ContinualUR,Wu2018MemoryRG}.  For instance, STAM \cite{stam} employed an expandable memory architecture designed for processing single-pass data streams by incorporating  novelty detection and memory update, while LUMP \cite{lump} enhances memory retention by augmenting data and blending new samples with stored representations.  These methods refine representations but rely on memory-intensive replay buffers or complex memory architectures, unsuitable in privacy-constrained or low-memory settings. CaSSLe \cite{cassle} transforms the self-supervised loss into a knowledge distillation approach by associating the present state of a representation with its preceding state. However, CaSSLe and similar methods only retain a single model from the previous task, which becomes inefficient when handling multiple tasks, leading to information loss over time. SCALE \cite{SCALE} addresses CF for non-iid and single-pass data using a self-supervised forgetting loss and online memory update mechanisms. POCON \cite{Gomez_Villa_2024_WACV} trains an expert network solely for new task, followed by an adaptation-retrospection phase to prevent forgetting. Evolve \cite{Yu_2024_WACV} leverages multiple pretrained models as cloud-based experts to enhance existing self-supervised learning methods on local clients.

In contrast, FBCC directly addresses catastrophic forgetting without replay by employing a forward–backward knowledge distillation mechanism. Unlike conventional KDs, which are typically for supervised transfer or compression with prior-data access, FBCC applies KD bi-directionally with multiple lightweight students that serve as task-specific memories, thereby extending KD to continual lifelong clustering without replay.

In parallel, several prototype-based continual learning methods have been proposed for supervised settings, where clustering is used as an auxiliary mechanism to summarize labeled class representations and stabilize replay. For instance, authors in  \cite{Aghasanli2025} propose a prototype-based framework with a label-free replay buffer and cluster preservation loss to mitigate catastrophic forgetting in class- and domain-incremental learning. However, their approach relies on class labels during training via supervised contrastive objectives, and clustering serves to compactly represent class structure rather than as the primary learning objective. 
Similarly, unsupervised continual domain shift learning is studied in \cite{Sun2025UCDSL}, where labels are unavailable during continual adaptation but class semantics are established through supervised pretraining. Their approach maintains multiple prototypes per class to model intra-class distributional shifts and preserve classification performance. In contrast, our setting assumes no predefined semantic classes, and clusters must be discovered and preserved over time. Prototype-based approaches have also been explored in non-class, task-specific continual learning, such as ProtoDepth \cite{Rim2025ProtoDepth}, which addresses unsupervised continual depth completion under domain shift. In this setting, prototypes act as domain-adaptive feature modifiers for a frozen backbone to mitigate forgetting in a regression task, rather than representing semantic classes or latent clusters. Unlike FBCC, ProtoDepth does not perform representation learning for structure discovery and does not aim to discover or preserve cluster identity.

\textbf{Deep Clustering.} Traditional deep clustering techniques \cite{wang2017research,6976982,Wu2024TransformerAutoencoder} typically involve finding a suitable latent representation (e.g., autoencoder (AE)) for data samples and subsequently employing a clustering algorithm, such as k-means, to assign the latent representation of samples to different clusters. Incorporating representation learning and clustering losses has been shown to yield better performance than traditional deep clustering methods in several studies \cite{IDEC,IDECF,DCN}. For example, DEC \cite{DEC} utilizes an autoencoder (AE) to jointly assign samples to clusters and refine representations based on the cluster assignments. \cite{IDEC,IDECF} enhance the performance of DEC by incorporating the reconstruction loss of the autoencoder into the loss function introduced by DEC. \cite{DML} employs a general autoencoder for instances that are easily clustered along with separate AEs for difficult-to-cluster data to improve performance. 

In recent years, contrastive learning has gained significant attention in unsupervised representation learning \cite{Yang2025Combined,albelwi2022survey,SimCLR}. 
Many deep clustering methodologies have seamlessly integrated contrastive learning, resulting in substantial improvements in deep clustering performance. Contrastive Clustering CC \cite{CC} applies contrastive learning at instance and cluster levels, while \cite{C3} extends it by incorporating inter-sample relationships in latent space of CC.  \cite{dang2021doubly} devises contrastive loss functions from both sample and class perspectives, thereby fostering the learning of more discriminative representations. Twin 
Contrastive Learning (TCL) \cite{li2022twin} extends CC by jointly performing instance- and cluster-level contrastive learning, enabling efficient clustering of streaming data where successive batches contain samples from the same fixed set of clusters as in the first batch, with no new classes or clusters appearing in the evolving incoming data. Therefore, it cannot accommodate newly emerging classes, unlike our continual clustering setting. 


While deep clustering approaches have achieved state-of-the-art performance across various tasks, to the best of our knowledge, no clustering algorithm has been specifically designed to handle data streams, where data clusters arrive sequentially. To address this gap, we propose FBCC in this paper, a novel approach developed for clustering in streaming data environments. 
\begin{figure*}[t]
  \centering
  \includegraphics[width=0.99\linewidth]{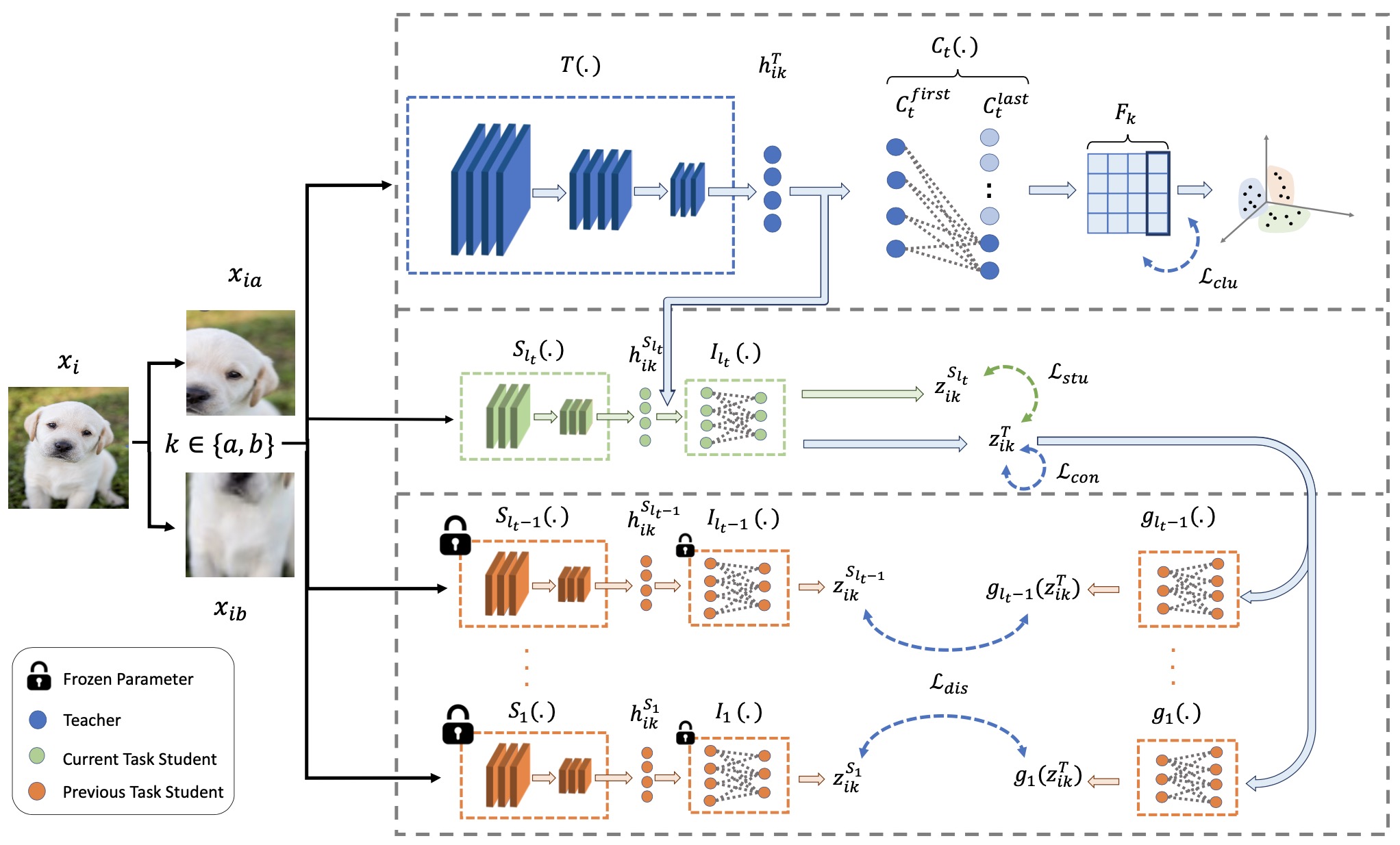}
\caption{Overview of the FBCC Framework for Task $t$: The teacher network, shown in blue, focuses on jointly clustering samples and learning representations for the current task while retaining knowledge of previous tasks with assistance from student networks trained on earlier tasks, depicted in orange. The student network for the current task, shown in green, aims to emulate the teacher network on the current task to assist the teacher in future tasks.}
\label{model}
\end{figure*}
\section{Method}
\label{sec:method}
We introduce the UCC problem which involves learning a sequence of $N$ tasks. Here, the set of tasks is denoted as $\Omega = \{\mathcal{D}_1, \mathcal{D}_2,...,\mathcal{D}_N\}$, where $\mathcal{D}_t$ corresponds to the dataset from task $t$ ($1\leq t \leq N$). In the UCC setting, although labeled information is not available, it is known that data samples belonging to different tasks are from distinct classes.  In other words, if $\mathcal{Y}_i$ and $\mathcal{Y}_j$ respectively represent sets of class labels from task $i$ and task $j$ ($i \neq j$), then $\mathcal{Y}_i \cap \mathcal{Y}_j = \varnothing$. This scenario is similar to unsupervised class-incremental setting \cite{cassle,wang2023comprehensive}; however, no class labels are available during training. Following the clustering literature where the number of clusters is predetermined, we assume that the number of clusters present in every task is known in the UCC setup, while cluster identities and assignments are discovered during training. In our benchmarks, the number of clusters per task is set equal to the number of underlying semantic classes for evaluation purposes only, and class labels are never used during training.

Given the recent success of contrastive clustering methods, we propose training FBCC's networks using losses inspired by contrastive learning.  In line with other contrastive learning methodologies, we employ two sets of augmentations ${a,b}$ on each sample within the dataset $\mathcal{D}_t =\{x_1,...,x_{|\mathcal{D}_t|}\}$ at task $t$, yielding $\mathcal{D}_{ta} =\{x_{1a},...,x_{|\mathcal{D}_t|a}\}$ and  $\mathcal{D}_{tb} =\{x_{1b},...,x_{|\mathcal{D}_t|b}\}$ respectively.

In the following sections, we describe our forward-backward knowledge distillation approach for continual clustering. Firstly, we introduce forward distillation, where the teacher is trained to learn the new task $t$ while leveraging knowledge from students 1 to $l_{t-1}$, whose parameters remain frozen. Here, $l_t$ is the index of the student model trained for task $t$. Conversely, in the backward mode, we propose a novel approach for training the $l_t$-th student to capture the teacher’s representation for task $t$. In the backward mode, all parameters of students ranging from 1 to $l_{t-1}$, as well as the teacher encoder, are frozen, and only the parameters of student $l_t$ are updated.
\subsection{Forward Knowledge Distillation}

\textbf{Training Teacher on the Current Task.} 
We train the teacher model to learn the current task, i.e. $\mathcal{D}_t$, while retaining knowledge from previous tasks. Since we do not have labeled data, we adopt contrastive learning \cite{CC,SimCLR} to guide the teacher's learning process. The model uses two augmentations per sample to produce similar representations while keeping them distinct from others in the batch. Each augmented view is generated by sequentially applying stochastic transformations including random resized cropping, horizontal flipping, color jittering, and random grayscale conversion to the same input sample, with full implementation details provided in the appendix. To further separate the current task’s representations from prior tasks, given the absence of datasets from previous tasks, we enforce distance using prototypes learned from previous tasks ranging from the first task up to task $t-1$ as $\mathcal{P}_{t-1}$. We discuss the way of defining prototypes later in this section.  This approach facilitates the preservation of task-specific information and aids in reducing interference between tasks, thereby enhancing the model's performance on sequential learning tasks. 
As shown in Figure~\ref{model}, the teacher encoder $T(\cdot)$ produces latent representations $h_{ik}^{T} = T(x_{ik})$, where $k \in \{a,b\}$ denotes different augmentations. To reduce dimensionality and compute contrastive losses, we use $l_t$ instance-level projectors $I_r(\cdot)$. For $1 \leq r \leq l_t - 1$, $I_r(\cdot)$ maps $h_{ik}^{S_r}$ from the $r$-th student to $z_{ik}^{S_r} = I_r(h_{ik}^{S_r})$. We propose to share the the $l_t$-th projector between the $l_t$-th student and the teacher, producing $z_{ik}^{S_{l_t}} = I_{l_t}(h_{ik}^{S_{l_t}})$ and $z_{ik}^{T} = I_{l_t}(h_{ik}^{T})$. 
The teacher encoder at task $t$ is trained using the following contrastive loss function (i.e. $\mathcal{L}_{con}$).

\begin{align}\label{eq_con}
    \mathcal{L}_{con} = \frac{1}{2|\mathcal{B}|} \sum_{x_i\in \mathcal{B}}(\ell_{ia}^{con} + \ell_{ib}^{con})
\end{align}
%
\begin{align}\label{lossa_con}
    &\ell_{ia}^{con}= -\log \nonumber\\
    &\scalebox{1}{$(\frac{\exp(sim(z_{ia}^{T},z_{ib}^{T}))}
    {\sum_{j \in \mathcal{B}}^{j\neq i}\sum_{k\in\{a,b\}}
    \exp(sim(z_{ia}^{T},z_{jk}^{T}))
    +\sum_{z_p \in \mathcal{P}_{t-1}}
    \exp(sim(z_{ia}^{T},z_p))})$}
\end{align}

In this study, $sim(.)$ denotes the cosine similarity between two vectors. Similar to $\ell_{ia}^{con}$, we define $\ell_{ib}^{con}$, which measures the similarity between $z_{ib}^{T}$ and other samples in the batch $\mathcal{B}$ as well as the prototypes set.

\textbf{Knowledge Distillation from Students to Teacher.}
In task $t$, since data from previous tasks are unavailable, we use student models trained on earlier tasks as memory to help the teacher retain prior knowledge and prevent forgetting. Each student specializes in one past task. Instead of directly matching $z_{ik}^{T}$ to $z_{ik}^{S_r}$, which could limit learning new concepts, we draw inspiration from \cite{cassle} and propose to map $z_{ik}^{T}$ using fully connected predictor networks $g_r(.)$ to the previous task learned by the $r$-th student (e.g. $z_{ik}^{S_r}$), while freezing the parameters of the $r$-th student and $r$-th instance-level projector. We apply a contrastive loss  as $\mathcal{L}_{dis}$ to align the predictor output with the latent representation of the previous task for the current dataset, as in \cite{cassle}. 

With this approach, our teacher network aims to imitate the behavior of students, which serve as estimations of the previous teacher. Unlike \cite{cassle}, which retains only the large-scale network from the last task, we store multiple lightweight students, each preserving knowledge of a specific task. This allows retention of multiple past tasks with far fewer parameters, reducing forgetting when many tasks are encountered. 

\textbf{Clustering Samples of the Current Task. }  We use a task-specific cluster-level projector $C_t(\cdot)$ to map teacher representations into a cluster probability space $F_k = [f_{1k} \mid \dots \mid f_{\lambda_t k}]$, where $k \in {a,b}$ denotes different augmentations, $\lambda_t$ is the number of clusters for task $t$, and each $f_{jk}$ is the cluster-assignment probability vector for cluster $j$. The first and last layers of $C_t(\cdot)$ are denoted as $C_t^{\text{first}}$ and $C_t^{\text{last}}$, respectively. We apply a cluster-level contrastive loss $\mathcal{L}_{clu}$, inspired by  \cite{CC}, to align probability assignments for similar clusters while separating dissimilar ones.  

\textbf{Updating Prototype Set.} At the end of training task $t$, we define $\lambda_t$ new prototypes that represent the clusters of task $t$. These prototypes, denoted as $p_v$ for $1 \leq v \leq \lambda_t$, are used in \eqref{eq_con} to ensure that representations of future tasks remain distinct from those learned during the current task. Once new prototypes have been identified for task $t$, we incorporate them into the existing set of prototypes from preceding tasks $P_{t-1}$ to form the updated prototype set $P_t$. 

Detailed formulations, initialization strategies and additional explanations for knowledge distillation from students to teacher, the task-specific clustering process and prototype computation are provided in the Appendix.
\subsection{Backward Knowledge Distillation}
In each task $t$, we train a light-weight student encoder ($l_t$-th student) with significantly fewer parameters than the teacher encoder to replicate its behavior. These student models are stored for knowledge distillation in future tasks, helping the teacher retain knowledge of clusters from previous tasks. However, in Section \ref{changing_m}, we demonstrate that considering all previous tasks when training the teacher for the current task is not essential. Therefore, we propose keeping up to $M$ trained lightweight student encoders, where $M$ is a hyperparameter (with $1 < M \le N$) that defines the maximum number of student models stored in memory at once. For $t \le M$, all previously trained students are retained and used for knowledge distillation in subsequent tasks. When $t > M$, only the most recent $M$ students are kept, and older ones are discarded. The optimal value of $M$ is determined by balancing memory usage and performance gain, as shown in Section \ref{changing_m}.  In our experimental evaluation on fixed benchmarks, $N$ is known only for analysis purposes to study the trade-off between performance and the number of retained students. Accordingly, $M=\lceil N/2 \rceil$ is selected as a representative operating point, since further increasing $M$ yields diminishing improvements in average accuracy and average forgetting while incurring additional memory overhead as discussed in Section \ref{changing_m}. The total parameters stored across these $M$ students remain smaller than that of the teacher encoder, providing better memory efficiency than methods such as \cite{cassle}, which store the full teacher network from the current task for knowledge distillation in subsequent tasks, as shown in Section 6 of the Appendix. The student is trained via $\mathcal{L}_{stu}$ to (1) match the teacher’s representations and (2) preserve the structural relationships between samples in a batch, enabling it to learn both output features and their inter-sample relations. Detailed loss formulations and explanations are provided in Appendix.

\subsection{Overall Training Scheme}
For each task $t$, FBCC alternates between forward distillation, where the teacher is trained using $\mathcal{L}_{con} + \mathcal{L}_{dis} + \mathcal{L}_{clu}$, and backward distillation, where the $l_t$-th student is trained using $\mathcal{L}_{stu}$. 

After completing training on the final task, the cluster assignments for all samples are obtained by leveraging the task-specific clustering spaces learned and stored during FBCC’s alternating forward and backward distillation, mapping each sample to the last layers trained for each task and selecting the index of the maximum value. Storing these task-specific cluster layers ensures that features from past tasks can still be mapped into their corresponding learned cluster spaces at test time, enabling accurate cluster assignment and performance evaluation without retraining. The pseudo-code for the FBCC training procedure (Algorithm 1) and the details of the final cluster assignment for samples are provided in the appendix.

\section{Experiments}
\label{sec:experiments}
In this section, we conduct comprehensive experiments to illustrate the effectiveness of our proposed method. We assess our model's performance on four challenging computer vision benchmark datasets by selecting certain classes as tasks in an incremental fashion: CIFAR-10 \cite{Cifar} (with 10 classes and 5 tasks), CIFAR-100 \cite{Cifar} (with 100 classes and 10 tasks), Tiny-ImageNet \cite{Imagenet} (with 200 classes and 10 tasks), and ImageNet100 \cite{Imagenet} (with 100 classes and 10 tasks). 
To train our model, we concatenate the train and test sets of the datasets, a common practice in clustering research (e.g., \cite{DCSS,DML,CC,C3}). 

As classes are disjoint across tasks, both the class identities and their sample proportions vary from task to task, resulting in a non-stationary data distribution. In contrast to experimental settings in which each task contains the same set of classes, yielding a stationary distribution and representing only new data rather than genuinely new tasks \cite{wang2024mostream}, our method explicitly models realistic scenarios where tasks arrive with different classes and varying class proportions.

\textbf{Implementation Details.} We use ResNet-18 \cite{resnet} as the teacher network, which has approximately 11.5 million parameters. For the student networks, we employ SqueezeNet 1.1 with around 1.2 million parameters. To match the output dimension of SqueezeNet to ResNet-18, we add a single-layer fully connected network at the end of SqueezeNet. 
 Instance projectors and predictors are 2-layer fully connected networks with dimensions $d\xrightarrow[]{}512\xrightarrow[]{}128$ ($d=512$ for projectors, $d=128$ for predictors). The cluster projector is a 2-layer network with dimensions $512\xrightarrow[]{}512\xrightarrow[]{}\lambda_t$, where $\lambda_t$ is the number of clusters in the task. The first layer of the cluster projector is shared, and the last layer is stored for cluster prediction. Batch size is 256 in all experiments.

\textbf{Metrics.} We evaluate the performance of our clustering model using two key metrics: average clustering accuracy ($\overline{\text{ACC}}$) and average forgetting ($\overline{\text{F}}$), where ACC 
is a widely used metric for assessing clustering performance and average forgetting is a common metric used to measure how much information the model has forgotten about previous tasks. $\overline{\text{ACC}}$ and $\overline{\text{F}}$ are defined as follows:
\begin{align}
    &\overline{\text{ACC}} = \frac{1}{\text{N}} \sum_{t=1}^{\text{N}} \text{ACC}_{t,\text{N}}\\
    \overline{\text{F}} &= \frac{1}{\text{N}-1} \sum_{i=1}^{\text{N}-1} \max_{t \in \{1,.., \text{N}-1\}} (\text{ACC}_{i,t}-\text{ACC}_{i,\text{N}}),
\end{align}
\\
where $\text{ACC}_{i,j}$ is $\text{ACC}$ of task $i$ at the end of training of task $j$.  

\begin{table}[htbp]
  \caption{FBCC performance comparison in terms of $\overline{\text{ACC}} (\%)$ and $\overline{\text{F}} (\%)$. }
    \centering
    \setlength{\tabcolsep}{0.5pt}
    \renewcommand{\arraystretch}{1.2}
    \scalebox{0.65}{\begin{tabular}{|c|c|c|c|c|c|c|c|c|}
        \hline
        \multirow{2}{*}{Algorithms} &
         \multicolumn{2}{c|}{CIFAR-10} & \multicolumn{2}{c|}{CIFAR-100} & \multicolumn{2}{c|}{Tiny-ImageNet}  & \multicolumn{2}{c|}{ImageNet100}\\
        \cline{2-9}
       & $\overline{\text{ACC}}$ $(\uparrow)$ & $\overline{\text{F}}$ $(\downarrow)$ &$\overline{\text{ACC}}$ $(\uparrow)$ & $\overline{\text{F}}$ $(\downarrow)$ &$\overline{\text{ACC}}$ $(\uparrow)$ & $\overline{\text{F}}$ $(\downarrow)$  & $\overline{\text{ACC}}$ $(\uparrow)$ & $\overline{\text{F}}$ $(\downarrow)$ 
       \\
       \hline
       CC (offline) & 79.00 & - & 42.90 & - & 14.00 & - & 47.60 & --\\
        \hline \hline
        Co$^2$L (SCL) & 28.35 & 14.05 & 19.88 &  10.30 & 8.69 & 4.95 & 23.12 & 10.34\\
        OCD-Net (SCL) & 40.41 & 7.03 & 18.06 &  7.46 & 7.97 & 5.31 & 19.83 & 8.41\\
        CCL & 36.56 & 6.21 & 19.59& 8.51 & 8.21 & 4.68 & 24.03 & 9.33\\
        STAM & 39.61  &  5.15 & 25.34  & 6.25 & 9.21 & 4.26 & 28.76 & 8.53 \\
        LUMP & 56.43 & 12.76 & 19.53& 6.16 & 10.53 & 2.51 & 32.70 & 7.63\\
        CaSSLe & 40.56 & 3.28 & 36.67 & 3.92 & 17.45 & 2.69 & 41.53 & 5.29\\
        POCON & 57.20 & 3.59 &35.29 & 4.28 & 16.25 & 3.54 & 39.28 & 5.88 \\
        \hline \hline
        \textbf{FBCC} & \textbf{75.28 $\pm$ 0.81} & \textbf{2.29 $\pm$ 0.29} &   \textbf{38.73 $\pm$ 0.64} & \textbf{3.62 $\pm$ 0.21}& \textbf{18.36 $\pm$ 0.32} & \textbf{2.00 $\pm$ 0.27} & \textbf{43.28 $\pm$ 0.66} & \textbf{4.79 $\pm$ 0.48}\\
       \hline 
    \end{tabular}}
    \label{table1} \vspace{-10mm}
\end{table} 

\subsection{Comparison Results}\label{comparison}
To the best of our knowledge, no existing unsupervised continual learning algorithm is designed explicitly for the clustering task. Therefore, in this section, we compare our proposed FBCC algorithm with state-of-the-art UCL algorithms such as CCL \cite{ccl}, STAM \cite{stam}, LUMP \cite{lump}, CaSSLe \cite{cassle}, and POCON \cite{Gomez_Villa_2024_WACV}. Also, we compare our FBCC with two state-of-the art SCL method Co$^2$L \cite{cha2021co2l} and OCD-Net \cite{ocd_net} on four benchmark datasets. Note that Co$^2$L and OCD-Net are supervised methods that make use of data labels during their training phase. 
We follow a widely-used method in the field, as described in \cite{SCALE}, by using the spectral clustering algorithm on the latent representations learned by other algorithms. This allows us to compute and report $\overline{\text{ACC}}$ and $\overline{\text{F}}$ for these techniques. The comparison results are shown in Table \ref{table1}. For each dataset, we conducted 5 experiments on FBCC using different random initializations and reported the average results along with the confidence intervals in this table.
Moreover, we compare our FBCC with a baseline CC \cite{CC} algorithm. CC possesses the flexibility to define its loss function using any pair of samples from distinct clusters, rendering it more potent compared to FBCC, which, in each step, only has access to partial clusters. Consequently, we employ CC as a proxy upper bound for assessing the performance of our algorithm.
Note that the lower performance of CC on the Tiny-ImageNet compared to FBCC can be associated with the fact that CC is a memory-hungry algorithm, and running it for more than 256 samples per batch is practically impossible \cite{C3,CC} while the given 256 samples might not be enough for defining relationship between samples when dealing with numerous clusters like the case of Tiny-ImageNet. Yet, CC serves as a proper upper bound for datasets with a small number of classes, such as CIFAR-10, CIFAR-100 and ImageNet100. As illustrated in Table \ref{table1}, FBCC demonstrates notable superiority over alternative algorithms concerning both $\overline{\text{ACC}}$ and $\overline{\text{F}}$. Notably, FBCC surpasses Co$^2$L and OCD-Net, which learns latent representations of data in a supervised manner, and the state-of-the-art UCL algorithm, CaSSLe, while employing fewer parameters. Specifically, FBCC utilizes a total of 15.1m parameters for CIFAR-10 and 17.5m for CIFAR-100, Tiny-ImageNet and ImageNet100, whereas CaSSLe employs 29m parameters across all datasets. FBCC’s advantage lies in two aspects: first, it is specifically designed for clustering tasks, jointly learning data representations and cluster assignments, unlike other UCL methods that focus only on representation learning; second, it preserves knowledge from previous tasks through a small set of specialized student models that mimic the teacher’s task-specific representations.

\subsection{Ablation Study}
\label{ablation}
\textbf{Effectiveness of Prototypes in Forward Knowledge Distillation.} To demonstrate the effectiveness of including prototypes learned from the previous task (i.e., $\mathcal{P}_{t-1}$) during training in the forward distillation phase, we propose to exclude prototypes and focus solely on the contrastive loss, i.e., we remove the second term in the denominator of Equation (2). All other model configurations remain unchanged. This setup is labeled as FBCC w/o Pro in Table \ref{tb2}. Upon comparing the results obtained from FBCC and FBCC w/o Pro, it is evident that the inclusion of prototypes in forward distillation leads to improved $\overline{\text{ACC}}$ and $\overline{\text{F}}$ across various tasks. This enhancement is attributed to our model's ability to effectively distinguish between data from the current task and prototypes, which serve as representatives of previous tasks.

\textbf{Effectiveness of Students in Forward Knowledge Distillation.} In Table \ref{tb2}, we present a comparison of our proposed method with two alternative configurations in terms of $\overline{\text{ACC}}$ and $\overline{\text{F}}$. In one of these configurations, denoted as FBCC w/o KD, we exclude the knowledge distillation loss from students to the teacher (i.e., $\mathcal{L}_{dis}$) when updating the parameters of the teacher encoder. In the second configuration, inspired by \cite{cassle}, instead of training multiple students, we employ a strategy where we utilize a previously trained teacher model to mitigate catastrophic forgetting. We freeze the parameters of this copied network, and the knowledge distillation loss is defined in \cite{cassle}. This configuration is labeled as FBCC + CaSSLe in Table \ref{tb2}.

If we compare FBCC w/o KD with FBCC, we observe approximately a 5.88\% improvement in terms of $\overline{\text{ACC}}$ and 7.05\% improvement in terms of $\overline{\text{F}}$ across all datasets. This improvement is primarily attributed to the effectiveness of knowledge distillation from students to the teacher using $\mathcal{L}_{dis}$ in retaining knowledge from previous tasks.

Moreover, upon comparing results obtained from FBCC + CaSSLe with those from FBCC, we can conclude that the effectiveness of having multiple students lies in retaining knowledge from more than one previous task. It is worth noting that the number of parameters for FBCC on CIFAR-10, CIFAR-100, Tiny-ImageNet and ImageNet100 are 15.1m, 17.5m, 17.5m, and 17.5m, respectively, while the number of parameters for FBCC + CaSSLe for all datasets is 29m. Our FBCC achieves better results in terms of 3.58\% improvement in $\overline{\text{ACC}}$ and exhibits 2.03\% improvement in terms of $\overline{\text{F}}$ across all datasets despite having fewer parameters.


\begin{table*}[t]
  \caption{Ablation study of FBCC in terms of $\overline{\text{ACC}} (\%)$ and $\overline{\text{F}} (\%)$. }
    \centering
    \setlength{\tabcolsep}{4pt}
    \renewcommand{\arraystretch}{1.5}
    \scalebox{0.8}{
    \begin{tabular}{|c|c|c|c|c|c|c|c|c|}
        \hline
        \multirow{2}{*}{Algorithms} &
         \multicolumn{2}{c|}{CIFAR-10} & \multicolumn{2}{c|}{CIFAR-100} & \multicolumn{2}{c|}{Tiny-ImageNet} &
         \multicolumn{2}{c|}{ImageNet100}\\
        \cline{2-9}
       & $\overline{\text{ACC}}$ $(\uparrow)$ & $\overline{\text{F}}$ $(\downarrow)$ &$\overline{\text{ACC}}$ $(\uparrow)$ & $\overline{\text{F}}$ $(\downarrow)$ &$\overline{\text{ACC}}$ $(\uparrow)$ & $\overline{\text{F}}$ $(\downarrow)$
       &$\overline{\text{ACC}}$ $(\uparrow)$ & $\overline{\text{F}}$ $(\downarrow)$\\
        \hline
        FBCC w/o Pro & 75.00 & 2.19 & 37.61 & 4.10 & 17.91 & 2.37 & 42.47 & 5.15\\
        FBCC w/o KD & 67.54 & 9.21 &  32.47  & 13.31 & 14.28 &6.58 & 37.17 & 12.89\\
        FBCC + CaSSLe & 70.69  & 4.63 & 35.21  & 6.41& 15.28& 3.01 & 39.62 & 6.94\\
        \hline 
        \textbf{FBCC} & \textbf{75.28} & \textbf{2.29} &   \textbf{38.73} & \textbf{3.62}& \textbf{18.36} & \textbf{2.00} & \textbf{43.28} & \textbf{4.79}\\
       \hline 
    \end{tabular}}
  
    \label{tb2}
\end{table*}

\begin{figure*}[h]
    \centering
    \includegraphics[width=0.85\textwidth]{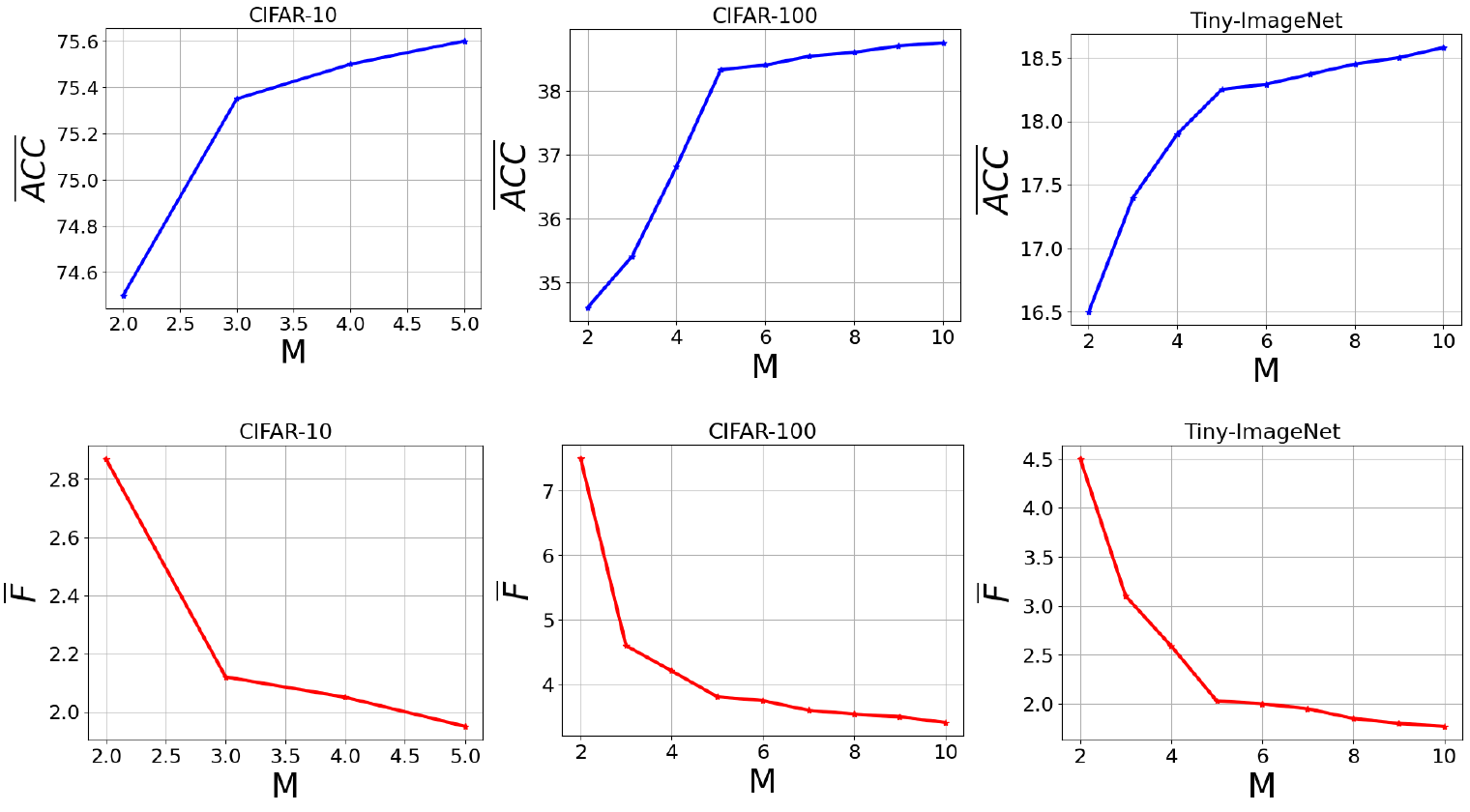}
    \caption{Average ACC and Average Forgetting of different values of $M$}
    \label{fig2}
\end{figure*}

\subsection{Effect of Number of Students in Forward Knowledge Distillation.} \label{changing_m}
In this section, we analyze the impact of the number of students (denoted as $M$), varied from 2 to $N$, on the performance of FBCC, measured by the average accuracy ($\overline{\text{ACC}}$) and average forgetting rate ($\overline{\text{F}}$), across CIFAR-10, CIFAR-100 and Tiny-ImageNet. The results are presented in Figure \ref{fig2}. As shown in the figure, a trade-off is observed between the performance of FBCC and the memory usage required for training the FBCC model. When the number of students increases, the model's ability to retain knowledge from previous tasks improves, albeit at the cost of higher memory consumption. 

As illustrated in Figure \ref{fig2}, an elbow-shaped trend emerges in the performance curve. Specifically, increasing $M$ from 2 to $\lceil \frac{N}{2}\rceil$ results in a notable performance improvement in terms of  $\overline{\text{ACC}}$ and $\overline{\text{F}}$, which can be attributed to the model's enhanced ability to retain information from earlier tasks. However, when $M$ is further increased from $\lceil \frac{N}{2}\rceil$ to $N$, the performance gains become marginal. This may be due to the increased number of students placing a greater burden on the teacher to retain knowledge from too many previous tasks, potentially reducing its ability to effectively learn new tasks. 

Based on these observations, we adopted $M=\lceil \frac{N}{2}\rceil$. This value captures most of the benefit of larger $M$ while avoiding the memory and compute overhead associated with storing and distilling from too many student models. Therefore, the $N$ is set to 3 for CIFAR-10, and 5 for CIFAR-100, Tiny-ImageNet, and ImageNet-100.

\subsection{Effect of Teacher Architecture}
In this section, we analyze the impact of various teacher architectures on the CIFAR-100 dataset, focusing on the performance of ResNet-18, ResNet-34, and ResNet-50 as teacher networks. The results, summarized in Table \ref{table4}, demonstrate that deeper architectures, such as ResNet-34 and ResNet-50, outperform ResNet-18. This improvement can be attributed to their enhanced capacity for capturing complex patterns and relationships within the data, leading to superior feature extraction. However, for consistency and fair comparison with other methods including \cite{cassle}, \cite{Gomez_Villa_2024_WACV}, \cite{ccl}, and \cite{lump} that use ResNet-18 as the continual learner, we also employ ResNet-18 in our experiments. This confirms that the success of FBCC is not simply due to using a more sophisticated teacher network.

\begin{table}[htbp]
  \caption{ Effect of different teacher architecture in terms of $\overline{\text{ACC}} (\%)$ and $\overline{\text{F}} (\%)$. }
    \centering
    \setlength{\tabcolsep}{4pt}
    \renewcommand{\arraystretch}{1.5}
   \scalebox{0.8}{ \begin{tabular}{|c|c|c|c|c|c|c|c|}
        \hline
        \multirow{2}{*}{Algorithms} &
         \multicolumn{2}{c|}{CIFAR-100}  \\
        \cline{2-3}
       & $\overline{\text{ACC}}$ $(\uparrow)$ & $\overline{\text{F}}$ $(\downarrow)$  \\
        \hline
        ResNet-18 & 38.73 & 3.62 \\
        
        ResNet-34 & 39.91 & 3.29\\
        ResNet-50 & 41.06 & 3.07\\
        \hline
    \end{tabular}}
  
    \label{table4}
\end{table}

Additional analyses are provided in the appendix, including:
(i) the effectiveness of FBCC in heterogeneous tasks, 
(ii) the impact of backward distillation when tested with three different student architectures, 
(iii) the efficacy of FBCC in tackling tasks with highly imbalanced sample distributions
, (iv) semantically proximal class groupings, e.g., CIFAR-100 super-class–based task definitions
, and (v) an efficiency comparison of FBCC against existing UCL benchmarks 
.

\section{Conclusion}
\label{sec:conclusion}
In conclusion, while UCL shows promise for sequential learning without labels, the lack of unsupervised continual clustering methods remains a challenge, particularly due to catastrophic forgetting (CF). Existing CF solutions such as knowledge distillation and replay buffers have limitations. To address this, we propose FBCC, which combines a continual teacher network with lightweight task-specific students. Through forward and backward distillation, FBCC enables incremental cluster discovery while preserving previously learned structure in a memory-efficient manner. Experiments on four benchmark datasets demonstrate that FBCC consistently outperforms existing UCL and SCL methods in both clustering accuracy and forgetting. These results highlight FBCC as a practical and scalable solution for continual clustering in dynamic, streaming environments.

\begin{credits}
\subsubsection{\ackname}

This work was supported by the Natural Sciences and Engineering Research Council of Canada (NSERC) and the Fonds de recherche du Québec - Nature et technologies (FRQNT).

\subsubsection{\discintname}
The authors declare that they have no known competing financial interests or personal relationships that could have appeared to influence the work reported in this paper.

\subsubsection{Declaration of Generative AI in the writing process.}
Generative AI tools were used to assist with formatting and language refinement. The authors reviewed and edited the output and take full responsibility for the final manuscript.

\end{credits}

%
%

%
%
%
%

\bibliographystyle{splncs04}
\bibliography{sec/main}

@String(CVPR= {IEEE Conf. Comput. Vis. Pattern Recog.})

@String(ICCV= {Int. Conf. Comput. Vis.})

@String(ECCV= {Eur. Conf. Comput. Vis.})

@String(NIPS= {Adv. Neural Inform. Process. Syst.})

@String(BMVC= {Brit. Mach. Vis. Conf.})

@String(ICME = {Int. Conf. Multimedia and Expo})

@String(ICIP = {IEEE Int. Conf. Image Process.})

@String(ICLR = {Int. Conf. Learn. Represent.})

@String(IJCAI = {IJCAI})

@String(AAAI = {AAAI})

@String(CVPRW= {IEEE Conf. Comput. Vis. Pattern Recog. Worksh.})

@String(CVPR  = {CVPR})

@String(ICCV  = {ICCV})

@String(ECCV  = {ECCV})

@String(NIPS  = {NeurIPS})

@String(BMVC  =	{BMVC})

@String(ICME  =	{ICME})

@String(ICIP  = {ICIP})

@String(ICLR  = {ICLR})

@String(CVPRW= {CVPRW})

@article{SimCLR,
  title={A Simple Framework for Contrastive Learning of Visual Representations},
  author={Chen, Ting and Kornblith, Simon and Norouzi, Mohammad and Hinton, Geoffrey},
  journal={arXiv preprint arXiv:2002.05709},
  year={2020}
}

@inproceedings{CC,
  author    = {Li, Y. and Hu, P. and Liu, Z. and Peng, D. and Zhou, J. T. and Peng, X.},
  title     = {Contrastive Clustering},
  booktitle = {Proceedings of the AAAI Conference on Artificial Intelligence},
  volume    = {35},
  pages     = {8547--8555},
  year      = {2021}
}

@inproceedings{DEC,
author = {Xie, Junyuan and Girshick, Ross and Farhadi, Ali},
title = {Unsupervised Deep Embedding for Clustering Analysis},
year = {2016},
publisher = {JMLR.org},
booktitle = {Proceedings of the 33rd International Conference on International Conference on Machine Learning - Volume 48},
pages = {478–487},
numpages = {10},
location = {New York, NY, USA},
series = {ICML'16}
}

@inproceedings{DCN,
  author    = {Yang, B. and Fu, X. and Sidiropoulos, N. D. and Hong, M.},
  title     = {Towards K-Means-Friendly Spaces: Simultaneous Deep Learning and Clustering},
  booktitle = {Proceedings of the 34th International Conference on Machine Learning (ICML)},
  pages     = {3861--3870},
  year      = {2017}
}

@inproceedings{IDEC,
author = {Guo, Xifeng and Gao, Long and Liu, Xinwang and Yin, Jianping},
title = {Improved Deep Embedded Clustering with Local Structure Preservation},
year = {2017},
isbn = {9780999241103},
publisher = {AAAI Press},
booktitle = {Proceedings of the 26th International Joint Conference on Artificial Intelligence},
pages = {1753–1759},
numpages = {7},
location = {Melbourne, Australia},
series = {IJCAI'17}
}

@misc{Cifar,
title= {CIFAR-10 (Canadian Institute for Advanced Research)},
journal= {},
author= {Alex Krizhevsky and Vinod Nair and Geoffrey Hinton},
year= {},
url= {http://www.cs.toronto.edu/~kriz/cifar.html},
terms= {}
}

@misc{Imagenet,
title= {Imagenet Full (Fall 2011 release)},
journal= {},
author= {Jia Deng and Wei Dong and Richard Socher and Li-Jia Li and Kai Li and Li Fei-Fei},
year= {},
url= {},
license= {},
tos= {},
superseded= {},
terms= {You have been granted access for non-commercial research/educational use. By accessing the data, you have agreed to the following terms.

You (the "Researcher") have requested permission to use the ImageNet database (the "Database") at Princeton University and Stanford University. In exchange for such permission, Researcher hereby agrees to the following terms and conditions:

1. Researcher shall use the Database only for non-commercial research and educational purposes.
2. Princeton University and Stanford University make no representations or warranties regarding the Database, including but not limited to warranties of non-infringement or fitness for a particular purpose.
3. Researcher accepts full responsibility for his or her use of the Database and shall defend and indemnify Princeton University and Stanford University, including their employees, Trustees, officers and agents, against any and all claims arising from Researcher's use of the Database, including but not limited to Researcher's use of any copies of copyrighted images that he or she may create from the Database.
4. Researcher may provide research associates and colleagues with access to the Database provided that they first agree to be bound by these terms and conditions.
5. Princeton University and Stanford University reserve the right to terminate Researcher's access to the Database at any time.
6. If Researcher is employed by a for-profit, commercial entity, Researcher's employer shall also be bound by these terms and conditions, and  Researcher hereby represents that he or she is fully authorized to enter into this agreement on behalf of such employer.
7. The law of the State of New Jersey shall apply to all disputes under this agreement.}
}

@inproceedings{resnet,
  title={Deep residual learning for image recognition},
  author={He, Kaiming and Zhang, Xiangyu and Ren, Shaoqing and Sun, Jian},
  booktitle={Proceedings of the IEEE conference on computer vision and pattern recognition},
  pages={770--778},
  year={2016}
}

@INPROCEEDINGS{IDECF,  author={Sadeghi, Mohammadreza and Armanfard, Narges},  booktitle={2021 IEEE International Conference on Image Processing (ICIP)},   title={IDECF: Improved Deep Embedding Clustering With Deep Fuzzy Supervision},   year={2021},  volume={},  number={},  pages={1009-1013}}

@article{DCSS,
  author  = {Sadeghi, M. and Soleimani, S. and Armanfard, N.},
  title   = {Deep Clustering with Self-Supervision Using Pairwise Similarities},
  journal = {IEEE Access},
  year    = {2025}
}

@article{DML, 
  author  = {Sadeghi, M. and Armanfard, N.},
  title   = {Deep Multi-Representation Learning for Data Clustering},
  journal = {TechRxiv},
  year    = {2022}
  }

@inproceedings{cassle,
  title={Self-supervised models are continual learners},
  author={Fini, Enrico and Da Costa, Victor G Turrisi and Alameda-Pineda, Xavier and Ricci, Elisa and Alahari, Karteek and Mairal, Julien},
  booktitle={Proceedings of the IEEE/CVF Conference on Computer Vision and Pattern Recognition},
  pages={9621--9630},
  year={2022}
}

@article{C3,
  title={C3: Cross-instance guided contrastive clustering},
  author={Sadeghi, Mohammadreza and Hojjati, Hadi and Armanfard, Narges},
  journal={The 34th British Machine Vision Conference (BMVC)},
  year={2023}
}

@article{acc,
  title={Image clustering using local discriminant models and global integration},
  author={Yang, Yi and Xu, Dong and Nie, Feiping and Yan, Shuicheng and Zhuang, Yueting},
  journal={IEEE Transactions on Image Processing},
  volume={19},
  number={10},
  pages={2761--2773},
  year={2010},
  publisher={IEEE}
}

@inproceedings{SCALE,
  title={SCALE: Online Self-Supervised Lifelong Learning without Prior Knowledge},
  author={Yu, Xiaofan and Guo, Yunhui and Gao, Sicun and Rosing, Tajana},
  booktitle={Proceedings of the IEEE/CVF Conference on Computer Vision and Pattern Recognition},
  pages={2483--2494},
  year={2023}
}

@article{squeezenet,
  title={SqueezeNet: AlexNet-level accuracy with 50x fewer parameters and< 0.5 MB model size},
  author={Iandola, Forrest N and Han, Song and Moskewicz, Matthew W and Ashraf, Khalid and Dally, William J and Keutzer, Kurt},
  journal={arXiv preprint arXiv:1602.07360},
  year={2016}
}

@inproceedings{ccl,
  title={Continual contrastive learning for image classification},
  author={Lin, Zhiwei and Wang, Yongtao and Lin, Hongxiang},
  booktitle={2022 IEEE International Conference on Multimedia and Expo (ICME)},
  pages={1--6},
  year={2022},
  organization={IEEE}
}

@article{lump,
  title={Representational continuity for unsupervised continual learning},
  author={Madaan, Divyam and Yoon, Jaehong and Li, Yuanchun and Liu, Yunxin and Hwang, Sung Ju},
  journal={arXiv preprint arXiv:2110.06976},
  year={2021}
}

@article{stam,
  title={Unsupervised progressive learning and the stam architecture},
  author={Smith, James and Taylor, Cameron and Baer, Seth and Dovrolis, Constantine},
  journal={Proceedings of the Thirtieth International Joint Conference on Artificial Intelligence},
  year={2021}
}

@article{alkhulaifi2021knowledge,
  title={Knowledge distillation in deep learning and its applications},
  author={Alkhulaifi, Abdolmaged and Alsahli, Fahad and Ahmad, Irfan},
  journal={PeerJ Computer Science},
  volume={7},
  pages={e474},
  year={2021},
  publisher={PeerJ Inc.}
}

@misc{rusu2022progressive,
      title={Progressive Neural Networks}, 
      author={Andrei A. Rusu and Neil C. Rabinowitz and Guillaume Desjardins and Hubert Soyer and James Kirkpatrick and Koray Kavukcuoglu and Razvan Pascanu and Raia Hadsell},
      year={2022},
      eprint={1606.04671},
      archivePrefix={arXiv},
      primaryClass={cs.LG}
}

@inproceedings{Rypes2024iclr,
title={Divide and not forget: Ensemble of  selectively trained experts  in Continual Learning},
author={Grzegorz Rype{\'s}{\'c} and Sebastian Cygert and Valeriya Khan and Tomasz Trzcinski and Bartosz Micha{\l} Zieli{\'n}ski and Bart{\l}omiej Twardowski},
booktitle={The Twelfth International Conference on Learning Representations},
year={2024}
}

@article{Wang_2023,
   author  = {Wang, L. and Zhang, X. and Li, Q. and Zhang, M. and Su, H. and Zhu, J. and Zhong, Y.},
  title   = {Incorporating Neuro-inspired Adaptability for Continual Learning in Artificial Intelligence},
  journal = {Nature Machine Intelligence},
  volume  = {5},
  number  = {12},
  pages   = {1356--1368},
  year    = {2023}
}

@inproceedings{Jung2020nips,
author = {Jung, Sangwon and Ahn, Hongjoon and Cha, Sungmin and Moon, Taesup},
title = {Continual learning with node-importance based adaptive group sparse regularization},
year = {2020},
isbn = {9781713829546},
publisher = {Curran Associates Inc.},
address = {Red Hook, NY, USA},
booktitle = {Proceedings of the 34th International Conference on Neural Information Processing Systems},
articleno = {308},
numpages = {12},
location = {Vancouver, BC, Canada},
series = {NIPS'20}
}

@inproceedings{Paik2019OvercomingCF,
  author    = {Paik, I. and Oh, S. and Kwak, T. and Kim, I.},
  title     = {Overcoming Catastrophic Forgetting by Neuron-Level Plasticity Control},
  booktitle = {Proceedings of the AAAI Conference on Artificial Intelligence},
  year      = {2019}
}

@InProceedings{Aljundi_2018_ECCV,
author = {Aljundi, Rahaf and Babiloni, Francesca and Elhoseiny, Mohamed and Rohrbach, Marcus and Tuytelaars, Tinne},
title = {Memory Aware Synapses: Learning what (not) to forget },
booktitle = {Proceedings of the European Conference on Computer Vision (ECCV)},
month = {September},
year = {2018}
}

@inproceedings{
arani2022learning,
  author    = {Arani, E. and Sarfraz, F. and Zonooz, B.},
  title     = {Learning Fast, Learning Slow: A General Continual Learning Method Based on Complementary Learning System},
  booktitle = {International Conference on Learning Representations (ICLR)},
  year      = {2022}
}

@inproceedings{yoon2022online,
  author    = {Yoon, J. and Madaan, D. and Yang, E. and Hwang, S. J.},
  title     = {Online Coreset Selection for Rehearsal-based Continual Learning},
  booktitle = {International Conference on Learning Representations (ICLR)},
  year      = {2022}
}

@article{RAMAPURAM2020381,
  author  = {Ramapuram, J. and Gregorova, M. and Kalousis, A.},
  title   = {Lifelong generative modeling},
  journal = {Neurocomputing},
  volume  = {404},
  pages   = {381--400},
  year    = {2020}
}

@inproceedings{Rao2019ContinualUR,
  author    = {Rao, D. and Visin, F. and Rusu, A. A. and Teh, Y. W. and Pascanu, R. and Hadsell, R.},
  title     = {Continual Unsupervised Representation Learning},
  booktitle = {Advances in Neural Information Processing Systems (NeurIPS 2019)},
  year      = {2019}
}

@inproceedings{Wu2018MemoryRG,
  author    = {Wu, C. and Herranz, L. and Liu, X. and Wang, Y. and van de Weijer, J. and Raducanu, B.},
  title     = {Memory Replay GANs: Learning to Generate Images from New Categories Without Forgetting},
  booktitle = {Advances in Neural Information Processing Systems (NeurIPS 2018)},
  year      = {2018}
}

@InProceedings{Gomez_Villa_2024_WACV,
     author    = {Gomez-Villa, A. and Twardowski, B. and Wang, K. and van de Weijer, J.},
  title     = {Plasticity-Optimized Complementary Networks for Unsupervised Continual Learning},
  booktitle = {Proceedings of the IEEE/CVF Winter Conference on Applications of Computer Vision (WACV)},
  pages     = {1690--1700},
  year      = {2024}
}

@InProceedings{Yu_2024_WACV,
  author    = {Yu, X. and Rosing, T. and Guo, Y.},
  title     = {Evolve: Enhancing Unsupervised Continual Learning with Multiple Experts},
  booktitle = {Proceedings of the IEEE/CVF Winter Conference on Applications of Computer Vision (WACV)},
  pages     = {2366--2377},
  year      = {2024}
}

@article{Howard2019SearchingFM,
  author    = {Howard, A. G. and Sandler, M. and Chu, G. and Chen, L.-C. and Chen, B. and Tan, M. and Wang, W. and Zhu, Y. and Pang, R. and Vasudevan, V. and Le, Q. V. and Adam, H.},
  title     = {Searching for MobileNetV3},
  booktitle = {Proceedings of the IEEE/CVF International Conference on Computer Vision (ICCV)},
  pages     = {1314--1324},
  year      = {2019}
}

@inproceedings{ma2018shufflenet,
  title={Shufflenet v2: Practical guidelines for efficient cnn architecture design},
  author={Ma, Ningning and Zhang, Xiangyu and Zheng, Hai-Tao and Sun, Jian},
  booktitle={Proceedings of the European conference on computer vision (ECCV)},
  pages={116--131},
  year={2018}
}

@article{iandola2016squeezenet,
  title={SqueezeNet: AlexNet-level accuracy with 50x fewer parameters and< 0.5 MB model size},
  author={Iandola, Forrest N and Han, Song and Moskewicz, Matthew W and Ashraf, Khalid and Dally, William J and Keutzer, Kurt},
  journal={arXiv preprint arXiv:1602.07360},
  year={2016}
}

@inproceedings{li2022learning,
  title={Learning from students: Online contrastive distillation network for general continual learning},
  author={Li, Jin and Ji, Zhong and Wang, Gang and Wang, Qiang and Gao, Feng},
  booktitle={Proc. 31st Int. Joint Conf. Artif. Intell.},
  pages={3215--3221},
  year={2022}
}

@article{wang2023comprehensive,
  title={A comprehensive survey of continual learning: Theory, method and application},
  author={Wang, Liyuan and Zhang, Xingxing and Su, Hang and Zhu, Jun},
  journal={arXiv preprint arXiv:2302.00487},
  year={2023}
}

@article{dang2021doubly,
  title={Doubly contrastive deep clustering},
  author={Dang, Zhiyuan and Deng, Cheng and Yang, Xu and Huang, Heng},
  journal={arXiv preprint arXiv:2103.05484},
  year={2021}
}

@article{albelwi2022survey,
  title={Survey on self-supervised learning: auxiliary pretext tasks and contrastive learning methods in imaging},
  author={Albelwi, Saleh},
  journal={Entropy},
  volume={24},
  number={4},
  pages={551},
  year={2022},
  publisher={MDPI}
}

@inproceedings{wang2021ordisco,
  title={Ordisco: Effective and efficient usage of incremental unlabeled data for semi-supervised continual learning},
  author={Wang, Liyuan and Yang, Kuo and Li, Chongxuan and Hong, Lanqing and Li, Zhenguo and Zhu, Jun},
  booktitle={Proceedings of the IEEE/CVF Conference on Computer Vision and Pattern Recognition},
  pages={5383--5392},
  year={2021}
}

@inproceedings{mai2021supervised,
  title={Supervised contrastive replay: Revisiting the nearest class mean classifier in online class-incremental continual learning},
  author={Mai, Zheda and Li, Ruiwen and Kim, Hyunwoo and Sanner, Scott},
  booktitle={Proceedings of the IEEE/CVF Conference on Computer Vision and Pattern Recognition},
  pages={3589--3599},
  year={2021}
}

@inproceedings{wang2017research,
  title={Research on intrusion detection based on feature extraction of autoencoder and the improved k-means algorithm},
  author={Wang, Xingang and Wang, Linlin},
  booktitle={2017 10th International Symposium on Computational Intelligence and Design (ISCID)},
  volume={2},
  pages={352--356},
  year={2017},
  organization={IEEE}
}

@INPROCEEDINGS{6976982,
author={Huang, Peihao and Huang, Yan and Wang, Wei and Wang, Liang},  booktitle={2014 22nd International Conference on Pattern Recognition},   title={Deep Embedding Network for Clustering},   year={2014},  volume={},  number={},  pages={1532--1537}}

@inproceedings{cha2021co2l,
  title={Co2l: Contrastive continual learning},
  author={Cha, Hyuntak and Lee, Jaeho and Shin, Jinwoo},
  booktitle={Proceedings of the IEEE/CVF International conference on computer vision},
  pages={9516--9525},
  year={2021}
}

@inproceedings{ocd_net, 
title={Learning from students: Online contrastive distillation network for general continual learning},
  author={Li, Jin and Ji, Zhong and Wang, Gang and Wang, Qiang and Gao, Feng},
  booktitle={Proc. 31st Int. Joint Conf. Artif. Intell.},
  pages={3215--3221},
  year={2022}
}

@misc{beclr2024,
  author = {Poulakakis-Daktylidis, S. and Jamali-Rad, H.},
  title  = {BECLR: Batch Enhanced Contrastive Few-Shot Learning},
  year   = {2024},
  note   = {arXiv:2402.02444}
}

@inproceedings{wang2024mostream,
  author    = {Wang, Z. and Wang, X. and Zhang, S.},
  title     = {MOStream: A Modular and Self-Optimizing Data Stream Clustering Algorithm},
  booktitle = {IEEE International Conference on Data Mining (ICDM)},
  pages     = {500--509},
  year      = {2024}
  }

@article{li2022twin,
  title={Twin contrastive learning for online clustering},
  author={Li, Yixin and Yang, Meng and Peng, Danni and Li, Tao and Huang, Junzhou and Peng, Xi},
  journal={International Journal of Computer Vision},
  volume={130},
  number={9},
  pages={2205--2221},
  year={2022},
  publisher={Springer}
}

@article{ashfahani2022,
  author    = {A. Ashfahani and M. Pratama},
  title     = {{Unsupervised continual learning in streaming environments}},
  journal   = {IEEE Transactions on Neural Networks and Learning Systems},
  volume    = {34},
  number    = {12},
  pages     = {9992--10003},
  year      = {2022}
}

@inproceedings{Aghasanli2025,
  author    = {Aghasanli, Agil and Li, Yi and Angelov, Plamen},
  title     = {Prototype-Based Continual Learning with Label-free Replay Buffer and Cluster Preservation Loss},
  booktitle = {Proceedings of the IEEE/CVF Conference on Computer Vision and Pattern Recognition Workshops (CVPRW)},
  year      = {2025},
  pages     = {6545--6554}
}

@inproceedings{Sun2025UCDSL,
  author    = {Sun, Haopeng and Zhang, Yingwei and Xu, Lumin and Jin, Sheng and Luo, Ping and Qian, Chen and Liu, Wentao and Chen, Yiqiang},
  title     = {Unsupervised Continual Domain Shift Learning with Multi-Prototype Modeling},
  booktitle = {Proceedings of the IEEE/CVF Conference on Computer Vision and Pattern Recognition (CVPR)},
  year      = {2025},
  pages     = {10131--10141}
}

@inproceedings{Rim2025ProtoDepth,
  author    = {Rim, P. and Park, H. and Gangopadhyay, S. and Zeng, Z. and Chung, Y. and Wong, A.},
  title     = {ProtoDepth: Unsupervised Continual Depth Completion with Prototypes},
  booktitle = {Proceedings of the IEEE/CVF Conference on Computer Vision and Pattern Recognition (CVPR)},
  year      = {2025},
  pages     = {6304--6316}
}

@article{Wu2024TransformerAutoencoder,
  author  = {Wu, W. and Wang, W. and Jia, X. and Feng, X.},
  title   = {Transformer Autoencoder for K-means Efficient Clustering},
  journal = {Engineering Applications of Artificial Intelligence},
  volume  = {133},
  pages   = {108612},
  year    = {2024}
}

@article{Yang2025Combined,
  author  = {Yang, Z. and Li, K. and Huang, Z. and Xu, Z. and Zhu, X. and Xiao, Y.},
  title   = {A combined perspective self-supervised contrastive learning framework for human activity recognition integrating instance prediction and clustering},
  journal = {Engineering Applications of Artificial Intelligence},
  volume  = {162},
  pages   = {112317},
  year    = {2025}
}
\newpage
\appendix
\section{Implementation Details}\label{appA}
In Section 4.3 of the main manuscript, we demonstrate that considering all previous tasks in training the current teacher is not essential. We propose training up to $M$ light-weight student encoders, where $M$ is a hyperparameter with $1 < M \leq N$, to alleviate catastrophic forgetting. Specifically, when $t < M$, we train $t$ students, and when $t \geq M$, we retain the last $M$ students while removing all others from memory. Henceforth, we denote the number of students for task $t$ as $l_t = \mathbbm{1}\{t < M\} t + \mathbbm{1}\{t \geq M\} M$, where $\mathbbm{1}\{.\}$ is an indicator function. The $r$-th student encoder is denoted as $S_r(.)$, where $1\leq r \leq l_t$. The latent representations of the $r$-th student encoder for $x_{ik}$ is denoted as $h_{ik}^{S_r}=S_r(x_{ik})$. Hereafter, we utilize the notation $r$ to denote indices ranging from 1 to $l_t$, i.e. $1\leq r\leq l_t$. 
 
\subsection{Forward Knowledge Distillation}
\textbf{Knowledge Distillation from Students to Teacher.} Given the well-studied effectiveness of utilizing contrastive loss between the output of the predictor and the latent representation of the previous task for the current dataset in \cite{cassle}, we define our loss as follows:
\begin{align}
    \mathcal{L}_{dis} = \frac{1}{2|\mathcal{B}|} \sum_{x_i\in \mathcal{B}}(\ell_{ia}^{dis} + \ell_{ib}^{dis}),
\end{align}
\vspace{-12mm}

\begin{align}\label{l_dis}
    &\ell_{ia}^{dis}= \nonumber \\ &\scalebox{1}{$- \frac{1}{l_t-1}\sum_{1 \leq r<l_t}\log(\frac{\exp(sim(g_r(z_{ia}^{T}),\Delta(z_{ia}^{S_{r}})))}{\sum_{j \in \mathcal{B}}^{j\neq i}\sum_{k\in\{a,b\}}\exp(sim(g_r(z_{ia}^{T}),\Delta(z_{jk}^{S_r})))})$}
\end{align}

where $\Delta(.)$ denotes the detaching operation, in which we detach vectors from a network and do not have a backward path to this network from our loss. With this approach, our teacher network aims to imitate the behavior of students, which serve as estimations of the previous teacher. The primary distinction between our proposed distillation framework and \cite{cassle} lies in our approach to addressing the catastrophic forgetting issue. We train multiple light-weight students, each capable of serving as a reliable estimation of our teacher network, allowing us to remember more than one previous task while storing a lower number of parameters in memory. In contrast, \cite{cassle} relies solely on the large-scale deep network learned in the previous task, potentially leading to forgetting of the initial tasks when confronted with numerous tasks. Moreover, in supplementary material, we elucidate the significance of retaining memories of more than one previous task in mitigating the catastrophic forgetting issue. 
\\ \\
\textbf{Clustering Samples of the Current Task. }  To effectively cluster the data samples from the current task, we employ a method that combines the representation learning of the teacher model with clustering. The teacher model processes each sample and generates a representation that is passed through a clustering layer specifically designed for the task at hand. This clustering layer helps assign each data point to one of the predefined clusters. Inspired by \cite{CC}, for clustering samples belonging to $\mathcal{D}_t$, we propose to train a cluster-level projector network for task $t$, denoted by $C_t(\cdot)$. This network consists of a 2-layer fully connected network followed by a softmax function, which maps the latent representation of the teacher encoder to a suitable space designed for the clustering task. The first layer and the last layer of $C_t(\cdot)$ are denoted as $C_t^{\text{first}}$ and $C_t^{\text{last}}$, respectively, i.e.  $C_t = C_t^{\text{last}}(C_t^{\text{first}})$. At the beginning of task $t$, where $2\leq t \leq N$, we initialize $C_t^{\text{first}}$ with $C_{t-1}^{\text{first}}$.  This initialization is intended to retain the information from previous tasks during this stage. The output of $C_t^{\text{first}}$ for sample $x_{ik}$ is denoted by $\hat{h}_{ik} = C_t^{\text{first}} (T(x_{ik}))$. Also, we propose to store $C_{t-1}^{\text{last}}$ in memory for use during the test phase. Within $C_t^{\text{last}}$, we allocate $\lambda_t$ neurons to transform $\hat{h}_{ik}$ into a specialized space designed for the clustering of data samples within the current task. We initialize $C_t^{\text{last}}$ with random values. For instance, if we assume 2 clusters per task, in Figure 1 of the main manuscript, for the new task, we add two new neurons, shown in dark blue, to create $C_t^{\text{last}}$. Also, we store neurons of previous tasks shown in light blue in memory. 

In every task, for a batch of data $\mathcal{B}$, we create two augmentations of the batch to obtain $\mathcal{B}_k$, where $k \in \{a,b\}$. We then pass these two augmented batches to the teacher network and the cluster-level projector to obtain $F_k=C_{t}(T(\mathcal{B}_k))$, where $F_k = [f_{1k}| f_{2k}|...|f_{\lambda_tk}]\in \mathbbm{R}^{|\mathcal{B}|\times \lambda_t}$, and $f_{jk} \in \mathbbm{R}^{|\mathcal{B}|}$ represents the probability vector for assigning samples from $\mathcal{B}_k$ to cluster $j$. Inspired by \cite{CC}, we apply contrastive loss on the features of $F_k$ (e.g., $f_{jk}$) instead of applying the contrastive loss between samples. The motivation stems from the derivation of $F_a$ and $F_b$ from two augmentations of the same batch. Therefore, similar clusters represented in $F_a$ and $F_b$ (e.g. $f_{ia}$ and $f_{ib}$) are expected to possess matching probability assignments and ideally be situated far apart from dissimilar clusters. This strategy is designed to promote distinct and well-separated clusters, thereby improving the overall quality of the clustering process. The loss is defined as follow:
\vspace{-1.5mm}
\begin{align}
    \mathcal{L}^{clu} = \frac{1}{2\lambda_t}\sum_{i=1}^{\lambda_t}(\ell_{ia}^{clu} + \ell_{ib}^{clu}) - H(F),
\end{align}
\vspace{-4mm}
\begin{align}
    \ell_{ia}^{clu}= - \log(\frac{\exp(sim(f_{ia},f_{ib}))}{\sum_{j=1\, j\neq i}^{\lambda_t}\sum_{k\in\{a,b\}}\exp(sim(f_{ia},f_{jk})}), 
\end{align}
where $H(F) = \sum_{k \in \{a,b\}}\sum_{j=1}^{\lambda_t} -\mathcal{Q}(f_{jk})\log(\mathcal{Q}(f_{jk}))$ is entropy of cluster assignments probabilities, where $\mathcal{Q}(f_{jk}) = ||f_{jk}||_1/||F_k||_1$ and $||.||_1$ denotes the $\ell_1$ norm. We maximize the entropy to avoid the trivial solution of converging all assignments to one cluster. 
\\ \\
\textbf{Updating Prototype Set.} At the end of training of task $t$, we propose to define $\lambda_t$ new prototypes that are representatives of task $t$. These prototypes, denoted as $p_v$ for $1 \leq v \leq \lambda_t$, are intended for use in Equation (1) of the main manuscript where we generate representations of future tasks to be distinct from the representations learned during the current task. $p_v$ is designed to maintain the same distance with samples belonging to the $v$-th cluster. Given $c_{ia}$ and $c_{ib}$ as the cluster assignments of $x_{ia}$ and $x_{ib}$ (i.e. $c_{ia} = \text{argmax}[C_t(T(x_{ia})]$ and $c_{ib} = \text{argmax}[C_t(T(x_{ib})]$), we can formulate the following equation for determining $p_v$.

\begin{align}
    p_v = \frac{\sum_{x_i \in \mathcal{B} } \mathbbm{1}\{c_{ia} = v\, \, \text{and}\, \, c_{ib} = v\}(z_{ia}^T + z_{ib}^T)}{2\sum_{x_i \in \mathcal{B} } \mathbbm{1}\{c_{ia} = v\, \, \text{and}\, \, c_{ib} = v\}}
\end{align}
To enhance the quality of prototypes for the clusters of task t, the prototype of the $v$-th cluster is the center of the ``reliable'' augmented samples in the z space. We consider a sample as ``reliable'' if both of its augmentations are assigned to the same cluster. Once new prototypes have been identified for task $t$, we incorporate them into the existing set of prototypes obtained from preceding tasks $\mathcal{P}_{t-1}$ to constitute the updated prototype set $\mathcal{P}_t$.

\subsection{Backward Knowledge Distillation}
\vspace{-1.5mm}
In task $t$, we propose to train the $l_t$-th student network with the following loss function, while keeping the parameters of the teacher encoder and all other student encoders frozen.
\vspace{-3mm}
\begin{align}
    \mathcal{L}_{stu} = \frac{1}{2|\mathcal{B}|}\sum_{x_i\in \mathcal{B}} (\ell_{ia}^{stu} + \ell_{ib}^{stu})
\end{align}
\vspace{-10mm}

\begin{align}\label{loss_stu}
    \nonumber \ell_{ia}^{stu}&= \frac{1}{|h_{ia}^{S_{l_t}}|}|| h_{ia}^{S_{l_t}}-\Delta(h_{ia}^{T})||_2^2
    \\ &-\log(\frac{\exp(sim(z_{ia}^{S_{l_t}},\Delta(z_{ia}^{T})))}{\sum_{j \in \mathcal{B}}^{j\neq i}\sum_{k\in\{a,b\}}\exp(sim(z_{ia}^{S_{l_t}},\Delta(z_{jk}^{T})))})
\end{align}
%
%
where $||.||_2$ represents the $\ell_2$ norm and $|h_{ia}^{S_{l_t}}|$ shows the number of elements in $h_{ia}^{S_{l_t}}$. With this loss function, we aim to instruct our student network in two critical aspects: 1- Our student must grasp the representations produced by the teacher network irrespective of other samples. This is achieved through the first component of our loss. 2- Our student network must follow the same structural relationships established by the teacher encoder within a batch; we propose to enforce such behavior through defining a contrastive loss between the student and teacher network outputs, as is shown in the second term of the loss defined in \eqref{loss_stu}. While the first term in \eqref{loss_stu} aligns the intermediate representations $h_{ia}^{S_{l_t}}$ and $\Delta(h_{ia}^T)$, the mapping from $h$ to $z$ via the projector is nonlinear, and alignment in $h$-space does not guarantee optimal separation of clusters in $z$-space. Therefore, the second term explicitly enforces discriminative structure in $z$-space by pulling positive pairs closer and pushing apart negative pairs, thereby enhancing cluster separability. These two components of the loss function enable the $l_t$-th student to learn both the output and the relationships between samples produced by the teacher encoder.
 \\
\textbf{Overall Training Scheme.}
For the batch $\mathcal{B}$ comprising data samples from task $t$, we initially fix the parameters of all students and conduct forward distillation to minimize the combined losses of contrastive, distillation, and clustering (i.e., $\mathcal{L}_{con} + \mathcal{L}_{dis} + \mathcal{L}_{clu}$). Subsequently, we proceed with backward distillation, wherein we freeze the parameters of the teacher and unfreeze the parameters of the $l_t$-th student, minimizing $\mathcal{L}_{stu}$ for the same batch $\mathcal{B}$. Figure 1 of the main manuscript shows the overall training scheme for task $t$.  Furthermore, Algorithm 1 in the Supplementary Material file provides the pseudo-code for the training of FBCC.
\\
\textbf{Assigning Samples to Clusters:} For the UCC setting, after completing training on the last task $N$, to find the final cluster assignments for sample $x_i$, we propose to use the trained teacher encoder (i.e., $T$), the first layer of the cluster projector (i.e., $C_N^{\text{first}}$), which is shared among all tasks, and the last layers of the cluster projector, which are task-specific and stored in memory (i.e., $C_1^{\text{last}},..., C_N^{\text{last}}$). To assign cluster label $c_i$ to data sample $x_i$, we propose to obtain the latent representation of data $\hat{h}_i = C_N^{\text{first}}(T(x_i))$, then map $\hat{h}_i$ to different clustering spaces using the last layers trained for each task and pick the index of the maximum value, i.e., $c_i = \text{argmax}[C_1^{\text{last}}(\hat{h}_i),..., C_N^{\text{last}}(\hat{h}_i)]$. 

\section{Effectiveness of FBCC in Heterogeneous Tasks}\label{Heterogeneous}
In this section, we delve into the impact of heterogeneous tasks, where the number of clusters varies across tasks. The primary challenge posed by such heterogeneity for a continual learner lies in addressing catastrophic forgetting (CF), wherein the model must retain knowledge from tasks with a high number of clusters while adapting to subsequent tasks. To assess our model's ability to handle such scenarios, we delineate two cases using the CIFAR-100 dataset, aiming to evaluate its performance and compare it with the state-of-the-art UCL algorithm CaSSLe \cite{cassle}. In \textbf{Case 1}, we define the number of clusters as 50-10-10-10-10-10 for tasks 1 through 5, respectively, from left to right. Here, our model encounters half of the total clusters in the initial task before facing a consistent number of clusters across subsequent tasks. In \textbf{Case 2} (50-30-10-5-5), we present a more challenging setup where our model encounters a higher number of clusters initially, followed by tasks with progressively fewer clusters. This case emphasizes the significance of ensuring that the model retains knowledge from tasks with a greater number of clusters while adapting to those with fewer clusters. Comparison results are presented in Table \ref{table1}. In this section, we employ three students for conducting experiments. As depicted in Table \ref{table1}, the performance of FBCC surpasses that of CaSSLe. This superiority can be attributed to the fact that most forgetting occurs after task 1 and then task 2 for both Case 1 and Case 2. Our model exhibits the capability to remember previous tasks more effectively due to the utilization of multiple well-trained students. In contrast, CaSSLe relies on a single teacher for retaining knowledge from each preceding task, leading to a gradual forgetting phenomenon. This effect is particularly pronounced for tasks 1 and 2. Additionally, it is worth noting that the number of parameters in our model for this experiment is 15.1 million, whereas the number of parameters in CaSSLe is 23 million. This significant difference in parameter count underscores the efficiency of our approach in achieving competitive performance with fewer parameters.

\begin{table}[htbp]
  \caption{ FBCC and CaSSLe performance on heterogeneous tasks scenario in terms of $\overline{\text{ACC}} (\%)$ and $\overline{\text{F}} (\%)$. The best result for continual learning algorithms in each column is highlighted in bold.}
    \centering
    \setlength{\tabcolsep}{4pt}
    \renewcommand{\arraystretch}{1.5}
   \scalebox{0.8}{ \begin{tabular}{|c|c|c|c|c|c|c|c|}
        \hline
        \multirow{2}{*}{Algorithms} &
         \multicolumn{2}{c|}{Case 1} & \multicolumn{2}{c|}{Case 2} \\
        \cline{2-5}
       & $\overline{\text{ACC}}$ $(\uparrow)$ & $\overline{\text{F}}$ $(\downarrow)$ &$\overline{\text{ACC}}$ $(\uparrow)$ & $\overline{\text{F}}$ $(\downarrow)$ \\
        \hline
        CaSSLe & 48.74 & 7.89 & 45.61&  8.51 \\
        
        \textbf{FBCC} & \textbf{51.21} & \textbf{4.33} &   \textbf{48.9} & \textbf{4.81} \\
        \hline
    \end{tabular}}
  
    \label{table1}
\end{table}

 \section{Effectiveness of Backward Distillation} \label{BD}

In this section, we delve into the effectiveness of $\mathcal{L}_{stu}$ in transferring knowledge from teacher to student on CIFAR-100 dataset. To achieve this, we define an average over the difference between ACC of the teacher and the ACC of the student. This is defined as follows:
\begin{align}
    \hat{\text{ACC}} = \frac{1}{N} \sum_{t\in \{1,...,N\}} (\text{ACC}_{t,t}^T - \text{ACC}_{t,t}^{S_{l_t}})
\end{align}
where $\text{ACC}_{t,t}^T$ and $\text{ACC}_{t,t}^{S_{l_t}}$ represent the ACC of the teacher and the student on the task $t$ after completing training on the task $t$, respectively. To obtain $\text{ACC}_{t,t}^{S_{l_t}}$ for dataset $\mathcal{D}_t$, after completing training on $\mathcal{D}_t$, we take the output of the student, i.e. $h_{\mathcal{D}_t}^{S_{l_t}}$, and feed it into the cluster projector learned during forward distillation to obtain cluster assignments for the dataset, i.e. $c_{\mathcal{D}_t}= \text{argmax}[C_t(h_{\mathcal{D}_t}^{S_{l_t}})]$. Subsequently, we compare these assignments with the true cluster assignments to compute the $\text{ACC}_{t,t}^{S_{l_t}}$. 

Moreover, we consider three architectures for the student, namely MobileNetV3 Small \cite{Howard2019SearchingFM}, ShuffleNetV2 (0.5x) \cite{ma2018shufflenet}, and SqueezeNet 1.1 \cite{iandola2016squeezenet}. Inspired by \cite{alkhulaifi2021knowledge}, to select the best model for our student, we define a distillation score (DS) that takes into account the size and accuracy of the student relative to those of the teacher network. This score helps us identify the optimal model for the student. The formula for DS is defined as follows:
\begin{align}
    DS = \alpha (\frac{\# Param_S}{\# Param_T}) + (1-\alpha)(1-\frac{1}{N}\sum_{t = 1}^{N} \frac{\text{ACC}_{t,t}^{S_{l_t}}}{\text{ACC}_{t,t}^T}),
\end{align}
where $\# \text{Param}_S$ and $\# \text{Param}_T$ represent the number of parameters of the student and teacher network, respectively, and $\alpha \in [0,1]$ is a hyperparameter that highlights the importance of the first ratio over the second one.  Lower DS values indicate better models. In our experiments, $\alpha$ is set to 0.5 and the teacher network is ResNet-18 with 11.5 million parameters.

\begin{table}[htbp]
  \caption{Comparison of different student architectures in terms of number of parameters, $\hat{\text{ACC}} (\%)$ and DS. The best result in each column is highlighted in bold.}
  \centering
  \setlength{\tabcolsep}{4pt}
  \renewcommand{\arraystretch}{1.5}

  \scalebox{0.8}{
    \begin{tabular}{|c|c|c|c|}
      \hline
      \multirow{2}{*}{Students} &
      \multicolumn{3}{c|}{CIFAR-100} \\
      \cline{2-4}
      & $\#\text{Param}_S$ $(\downarrow)$
      & $\hat{\text{ACC}}$ $(\downarrow)$
      & $\text{DS}$ $(\downarrow)$ \\
      \hline
      MobileNetV3 Small & 2.5m & \textbf{1.05} & 0.120 \\
      ShuffleNetV2 (0.5x) & 1.3m & 1.91 & 0.081 \\
      SqueezeNet 1.1 & \textbf{1.2m} & 1.68 & \textbf{0.075} \\
      \hline
    \end{tabular}
  }

  \label{table2}
\end{table}

Table \ref{table2} shows the comparison of different student architecture in terms of number of parameters, $\hat{\text{ACC}} (\%)$, and DS. By comparing the $\hat{\text{ACC}}$ of students with different architectures, we can infer the effectiveness of $\mathcal{L}_{stu}$ in knowledge distillation from the teacher to the students. For instance, the ACC of SqueezeNet 1.1 with 1.2 million parameters is 1.68\% less than the ACC of ResNet-18 with 11.5 million parameters on average across different tasks on the CIFAR-100 dataset.

Based on DS reported in Table \ref{table2}, we choose SqueezeNet 1.1 as our student network for the CIFAR-100 dataset, as it has the lowest DS among the other architectures. We observe similar pattern for the other datasets as well.

\section{Imbalanced Dataset}\label{Imbalanced}

In this section, we delve into assessing the efficacy of our proposed FBCC in tackling learning tasks characterized by highly imbalanced sample distributions. To accomplish this, we adopt a strategy wherein we selectively sample data from task $t$ within the CIFAR-10 dataset. Instances from the first task are incorporated into the training set with a likelihood of 0.1, while instances from the final task are included with a likelihood of 1. Instances from intermediate tasks are chosen proportionally, following a linear progression. The inherent challenge posed by imbalanced data lies in the scenario where our model is trained on a limited number of instances from the current task, yet it encounters increasingly more samples from subsequent tasks. This imbalance heightens the risk of CF wherein the model's performance on the current task deteriorates as it learns new tasks, potentially leading to performance degradation in future tasks.
\begin{figure*}[t]
    \centering
    \includegraphics[width=1.0 \textwidth]{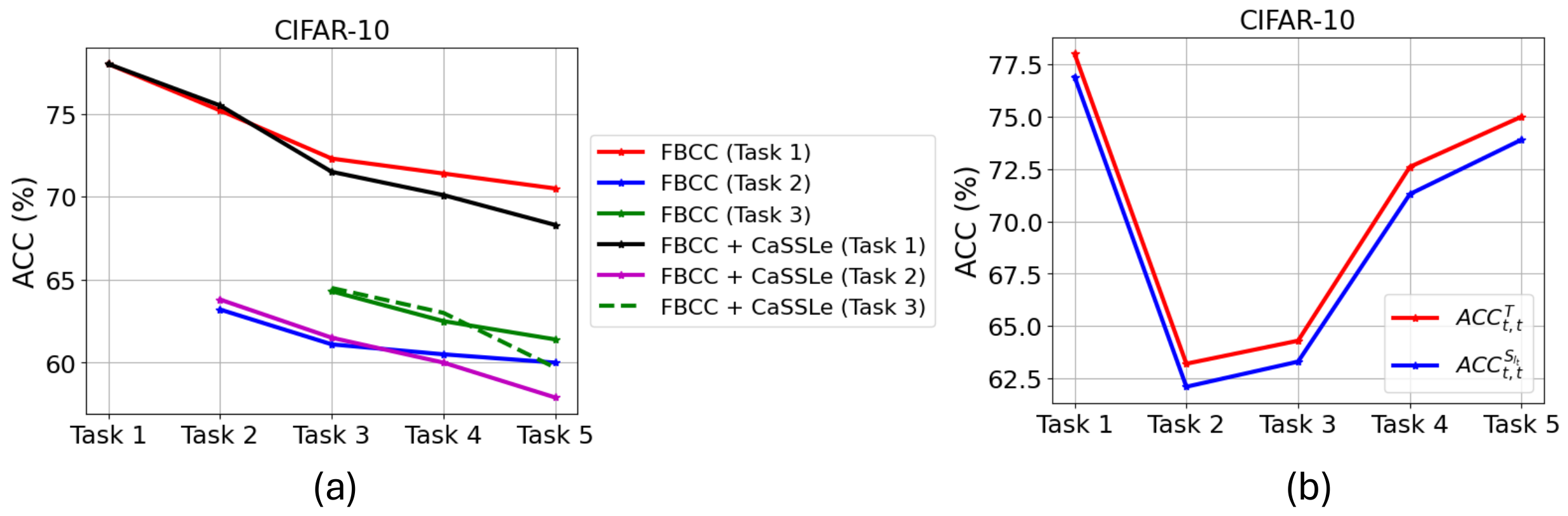} 
    \caption{Experiments on imbalanced data.\vspace{-4mm}}
    \label{fig3}
\end{figure*}

In Figure \ref{fig3}-(a), we present a comparative analysis between our proposed FBCC approach and FBCC + CaSSLe. The figure illustrates ACC achieved on the first three tasks ($1\leq t \leq 3$) after completing training on each task. As depicted in the figure, transitioning from task $t$ to task $t+1$ reveals that the performance of FBCC + CaSSLe for task $t$ surpasses that of FBCC. This discrepancy arises due to the utilization of a teacher network explicitly trained on task $t$ within the FBCC + CaSSLe framework. In contrast, FBCC employs a student network with notably fewer parameters for task retention. However, following task $t+2$, a reversal in performance is observed. FBCC exhibits superior performance on task $t$ compared to FBCC + CaSSLe. This shift can be attributed to FBCC's utilization of multiple specialized student networks, thereby enhancing its capacity to retain knowledge from previous tasks effectively. For example, this characteristic becomes particularly evident when examining the performance on task 1. At the conclusion of training on task 2, FBCC + CaSSLe exhibits superior performance compared to FBCC. However, after completing task 3, the trend reverses, with FBCC surpassing FBCC + CaSSLe in terms of ACC. 

Furthermore, to demonstrate the effectiveness of training students to mimic the behavior of the teacher network on imbalanced data, we plot $\text{ACC}_{t,t}^{\text{T}}$ and $\text{ACC}_{t,t}^{\text{S}_{l_t}}$, as discussed in Section 3 of the supplementary file, for imbalanced data in Figure \ref{fig3}-(b). For our experiments, we utilize SqueezeNet 1.1 \cite{squeezenet}. As depicted in the figure, our student network adeptly follows the teacher network in generating quality representations for the each task on imbalanced datasets.

\section{Semantic Proximity for Defining Tasks}\label{semantic}
CIFAR-100 consists of 100 fine-grained classes, which are further grouped into 20 broader super-classes. Each super-class represents a general category that includes multiple fine classes based on their similarities or related characteristics. For instance, the super-class ``vehicles 1" might include fine classes such as ``bicycle," ``bus," ``motorcycle," and others.

In this section, we introduce two different experimental settings for CIFAR-100. In the first setting, referred to as Case 1, we randomly select 10 tasks. This approach is commonly adopted in continual learning research \cite{cassle}, \cite{ccl}, and \cite{lump}. In the second setting, or Case 2, we ensure that each task contains data samples from two distinct super-classes within the CIFAR-100 dataset.

We apply our FBCC algorithm to both settings, with the results presented in Table \ref{table5}. As shown in the table, Case 2 outperforms Case 1, which can be attributed to the following factor: data samples within the same super-class tend to cluster closely together in the latent space, making it difficult for FBCC to distinguish between them. By presenting data samples from the same super-class as a task to the network, we enable the network to better differentiate between the various clusters within that super-class. In contrast, Case 1 relies primarily on prototypes to separate the super-classes, leading to less effective separation.
\begin{table}[htbp]
  \caption{Performance comparison of FBCC algorithm on different settings of CIFAR-100. The best result for continual learning algorithms in each column is highlighted in bold.}
    \centering
    \setlength{\tabcolsep}{4pt}
    \renewcommand{\arraystretch}{1.5}
   \scalebox{0.8}{ \begin{tabular}{|c|c|c|c|c|c|c|c|}
        \hline
        \multirow{2}{*}{Algorithms} &
         \multicolumn{2}{c|}{CIFAR-100}  \\
        \cline{2-3}
       & $\overline{\text{ACC}}$ $(\uparrow)$ & $\overline{\text{F}}$ $(\downarrow)$  \\
        \hline
        Case 1 & 38.73 & 3.62 \\
        
        Case 2 & \textbf{39.49} & \textbf{3.38}\\
     
        \hline
    \end{tabular}}
  
    \label{table5}
\end{table}

\begin{table*}[h]
  \caption{FBCC computational and memory efficiency comparison against existing UCL.}
  \centering
  \setlength{\tabcolsep}{3pt}
  \renewcommand{\arraystretch}{1.5}
  \scalebox{0.68}{
  \begin{tabular}{|c|c|c|c|c|}
    \hline
    Algorithms & Total Training Time (s) & Max GPU Memory (MB) & Model Size (MB) & Trainable Parameters (m) \\
    \hline
    CCL & 25133.49   & 1221.85 & 164.03 & 11.50 \\
    STAM & 1094.30 & 32650.25 & 14.23 & -- \\
    LUMP & 18856.38  & 2602.76 & 86.85 & 22.73  \\
    CaSSLe & 36810.84 & 2088.37 & 110.89 & 29.06  \\
    POCON & 23495.17 & 3920.75 & 402.67 & 49.52  \\
    \hline \hline 
    \textbf{FBCC} & 22591.10 & 19611.12 & 66.75 & 17.50  \\
    \hline
  \end{tabular}
  }
  \label{table6}
\end{table*}
\section{Efficiency Analysis of FBCC Against Existing UCL Benchmarks}\label{efficiency}

In addition to clustering performance, we assess the computational and memory efficiency of all models using CIFAR-100 dataset. Table \ref{table6} presents a comparison based on total training time, peak GPU memory usage, total model size, and the number of trainable parameters. Since our focus is on unsupervised continual learning (UCL), we restrict this efficiency analysis to models that specifically address this setting. STAM reports the lowest model size (14.23 MB) and fastest training time (1094 s) due to its minimalist architecture: it does not rely on gradient-based learning or neural weights and avoids backpropagation entirely but instead uses online clustering to store a set of certain centroids per task in a dual-memory system. Since these centroids are not trainable parameters in the conventional sense, we exclude this field in the table for STAM to avoid misrepresentation. Despite its efficiency, STAM performs poorly across all four datasets, specifically, on average, its clustering accuracy is over 30\% lower, and its forgetting scores are more than twice as high, reflecting limited retention and representation quality. For instance, on CIFAR-10, STAM reaches only 39.61\% $\overline{\text{ACC}}$ with an $\overline{\text{F}}$ of 5.15, while FBCC achieves 75.28\% $\overline{\text{ACC}}$ and 2.29 $\overline{\text{F}}$, clearly demonstrating the performance gap in both discriminative power and stability over time (see Table 1 of the main manuscript). This tradeoff highlights STAM’s limited capacity for capturing complex representations, which constrains its generalizability. Additionally, while STAM does not store raw data, its long-term memory accumulates task-specific centroids over time, resulting in a growing internal memory footprint. These centroids indirectly encode previously seen class information, which may raise concerns in memory-constrained or privacy-sensitive applications.
In contrast, our proposed FBCC model offers a significantly better balance compared to existing continual learning models: it achieves state-of-the-art performance across all datasets, with higher clustering accuracy and lower forgetting, while maintaining a moderate model size, reasonable training time and a manageable number of trainable parameters. Although FBCC uses 17.5 million trainable parameters which is slightly more than trainable parameters used by CCL, its overall model size is significantly smaller compared to CCL. This demonstrates FBCC’s practical memory efficiency despite its modular design. The increase in trainable parameters is justified by FBCC’s architecture, which includes multiple lightweight student networks and task-specific clustering heads. These components are essential for achieving superior clustering performance and robustness to forgetting. In contrast, CCL employs a single encoder trained end-to-end, limiting its flexibility and expressiveness in complex continual learning scenarios. Compared to larger models like POCON and CaSSLe, FBCC is more scalable, with lower computational overhead and higher performance, making it a compelling choice for continual clustering in both performance-critical and resource-constrained settings. Although FBCC exhibits higher peak GPU usage than some lighter models, this overhead stems solely from architectural complexity, namely, the simultaneous training of student and teacher networks and not from storing or replaying past data. Importantly, FBCC trades off replay buffer memory, which grows with the number of tasks, for temporary GPU allocation during training. This constitutes a one-time training resource cost, whereas replay-based methods impose continuous storage and sampling overhead across tasks. Unlike methods such as LUMP that rely on replay buffers, FBCC requires no access to previous task samples, making it privacy-compliant by design and better suited for settings where data retention is restricted. As a result, FBCC offers a practical and efficient solution for continual clustering that balances accuracy, scalability, and privacy.

\begin{algorithm}
\caption{Training of FBCC Algorithm for task $t$}

\begin{algorithmic}[1] \setstretch{1.1}
\Statex \textbf{Input:} Teacher network $T(\cdot)$, Student networks $S_1,..., S_{l_{t}}$ , Instance Projectors $I_1,... I_{l_t}$, Predictor networks $g_1,..., g_{l_{t}-1}$, prototype set $\mathcal{P}_{t-1}$, cluster-projector $C_t = C_t^{first}(C_t^{last})$, maximum number of iterations $Max_{iter}$, two sets of augmentations denoted by aug\_a and aug\_b, dataset for task $t$ shown by $\mathcal{D}_t$.
\Statex
\For{$1\leq epoch \leq Max_{iter}$}
\For{$\mathcal{B} = \{x_1,x_2,...x_{|\mathcal{B}|}\} \in \mathcal{D}_t$}
\State $\mathcal{B}_{a}$, $\mathcal{B}_{b}$ $\leftarrow$ aug\_a($\mathcal{B}$), aug\_b($\mathcal{B}$), where $\mathcal{B}_{k} = \{x_{ik}\}$ with $k\in\{a,b\}$ and $1 \leq i \leq |\mathcal{B}|$
\Statex
\Statex   \,\,\,\,\,\,\,\, \# \textbf{Forward Knowledge Distillation} 
\State Freeze Students and Unfreeze the Teacher
\State  $h_{ia}^{T}, h_{ib}^{T} \leftarrow T(x_{ia}), T(x_{ib})$ \Comment{we define  $h_k^T = \{h_{ik}^T\}$}
\State  $z_{ia}^{T}, z_{ib}^{T} \leftarrow I_{l_t}(h_{ia}^{T}), I_{l_t}(h_{ib}^{T})$ \Comment{we define  $z_k^T = \{z_{ik}^T\}$}
\State  $F_a, F_b \leftarrow C_t(T(\mathcal{B}_a)), C_t(T(\mathcal{B}_b))$
\For{$1\leq r \leq l_t-1$}
\State  $h_{ia}^{S_r}, h_{ib}^{S_r} \leftarrow S_r(x_{ia}), S_r(x_{ib})$ 
\State  $z_{ia}^{S_r}, z_{ib}^{S_r} \leftarrow I_{r}(h_{ia}^{S_r}), I_{r}(h_{ib}^{S_r})$ \Comment{we define  $z_k^{S_r} = \{z_{ik}^{S_r}\}$}
\EndFor
\State Compute $\mathcal{L}_{con}(z_{a}^{T}, z_{b}^{T}, \mathcal{P}_{t-1})$ using eq. (1) 
\State Compute $\mathcal{L}_{dis}(g_r(z_{k}^{T}), z_{k}^{S_r}.detach())$ using eq. (5) \Comment{$1\leq r\leq l_t-1$} 
\State Compute $\mathcal{L}_{clu}(F_a, F_b)$ using eq. (7)

\State Update parameters of $T$, $I_{l_t}$, $C_t$, and $g_r$ using $\mathcal{L}_{con} +\mathcal{L}_{clu} + \mathcal{L}_{dis}$ 
\Statex   
\Statex   \,\,\,\,
\textbf{\# Backward Knowledge Distillation} 
\State Freeze The Teacher and Unfreeze $l_t$-th Student
\State  $h_{ia}^{S_{l_t}}, h_{ib}^{S_{l_t}} \leftarrow S_{l_t}(x_{ia}), S_{l_t}(x_{ib})$ \Comment{we define  $h_k^{S_r} = \{h_{ik}^{S_r}\}$}
\State  $z_{ia}^{S_{l_t}}, z_{ib}^{S_{l_t}} \leftarrow I_{l_t}(h_{ia}^{S_{l_t}}), I_{l_t}(h_{ib}^{S_{l_t}})$ \Comment{we define  $z_k^{S_{l_t}} = \{z_{ik}^{S_{l_t}}\}$}
\State Compute $\mathcal{L}_{stu}(h_k^{S_{l_t}}, z_k^{S_{l_t}}, h_k^T.detach(),z_k^{T}.detach())$ using eq. (11)
\State Update parameters of $S_{l_t}$ using $\mathcal{L}_{stu}$ 
\EndFor
\EndFor
\State Update Prototype set $\mathcal{P}_{t-1}$ using (9) to obtain $\mathcal{P}_t$
\end{algorithmic}
\end{algorithm}
\newpage

\end{document}